\pdfoutput=1

\documentclass[11pt]{article}

\usepackage{ACL2023}

\usepackage{times}
\usepackage{latexsym}
\usepackage{tabularray}
\usepackage{amsmath}
\usepackage{multirow}
\usepackage{multicol}
\usepackage{booktabs}
\usepackage{makecell}
\usepackage{graphicx}
\usepackage{enumerate}
\usepackage{hhline}

\usepackage[ruled,vlined]{algorithm2e}

\usepackage[T1]{fontenc}

\usepackage[utf8]{inputenc}

\usepackage{microtype}

\usepackage{inconsolata}

\usepackage{hyperref}
\hypersetup{
    colorlinks=true,
    linkcolor=blue,
    filecolor=magenta,      
    urlcolor=cyan,
    pdftitle={Overleaf Example},
    pdfpagemode=FullScreen,
    }

\usepackage{tcolorbox} 
\usepackage{enumitem}   
\usepackage{pifont}     
\usepackage{setspace}   

\usepackage[draft,textsize=footnotesize,textwidth=15mm]{todonotes}

\usepackage[draft,textsize=footnotesize,textwidth=15mm]{todonotes}

\title{From Prejudice to Parity: A New Approach to Debiasing Large Language Model Word Embeddings}

\author{
\textbf{Aishik Rakshit}\\
\small Indian Institute of Technology, Guwahati\\
\And
\textbf{Smriti Singh}\\
\small University of Texas at Austin\\
\And
\textbf{Shuvam Keshari}\\
\small University of Texas at Austin\\
\AND
\textbf{Arijit Ghosh Chowdhury}\textsuperscript{\textdagger}\\
\small Amazon GenAI\\
\And
\textbf{Vinija Jain}\thanks{\,\,\,Work does not relate to position at Meta.}\\
\small Meta AI\\
\And
\textbf{Aman Chadha}\thanks{\,\,\,Work does not relate to position at Amazon.}\\
\small Amazon GenAI\\
}

\begin{document}
\maketitle
\begin{abstract}
Embeddings play a pivotal role in the efficacy of large language models. They are the bedrock on which these models grasp contextual relationships and foster a more nuanced understanding of language and consequently perform complex tasks that require a fundamental understanding of human language. Given that these embeddings themselves often reflect or exhibit bias, it stands to reason that these models may also inadvertently learn this bias. In this work, we build on the aforementioned seminal work of \citet{bolukbasi2016man} and \cite{gonen2019lipstick} and propose \textit{DeepSoftDebias}, an algorithm that uses a neural network to perform `soft debiasing'. We exhaustively evaluate this algorithm across a variety of state-of-the-art datasets, accuracy metrics, and challenging NLP tasks. On a wide range of metrics, we find that \textit{DeepSoftDebias} outperforms the current state-of-the-art methods at reducing bias across gender, race, and religion. 
\end{abstract}






\section{Introduction}

Word embeddings are a foundational element in the architecture of Large Language Models (LLMs). They act as the basis for these models to understand and subsequently, generate human-like language. However, it has been shown that these word embeddings may reflect or exhibit bias \cite{dev2020measuring,may2019measuring,Caliskan_2017}. Given the exponential increase in the use of LLMs on a plethora of downstream tasks, these representations can amplify bias and result in discriminatory actions, especially when it comes to the fields of education, healthcare, and justice. Existing work in this field has looked most commonly into gender bias \cite{kotek2023gender, bordia2019identifying,de2021stereotype}, racial bias \cite{mozafari2020hate,omiye2023large,tanglarge}, and religious bias \cite{baligudam2022systematic,kirk2021bias}. In this work, we build on the seminal work of \cite{gonen2019lipstick}, which brought attention to the inherent biases present in traditional GloVe embeddings \cite{pennington2014glove}. This study prompted the NLP community to re-evaluate the fundamental choices underlying our word representation models. Specifically, we present \textit{DeepSoftBias}: an algorithm that furthers the application of their methodology, by diverging from the conventional GloVe embeddings and delving into the word embeddings produced by the best-performing models on the Massive Text Embedding Benchmark (MTEB) \cite{muennighoff2022mteb} leaderboard. By employing these advanced embeddings on the same set of words as used in GloVe embeddings, we seek to investigate whether these state-of-the-art (SoTA) models inherently exhibit reduced bias. 

Our primary objective is two fold: first, to de-bias the embeddings from these selected models, and second, to rigorously assess the effectiveness of the bias removal process. Our proposed approach, \textit{DeepSoftDebias}, is an innovative methodology to de-bias LLM word embeddings which involves integrating a neural network into the soft debiasing approach developed by \citet{bolukbasi2016man}. This novel amalgamation is driven by the aspiration to enhance the debiasing process and contribute to the ongoing discourse on creating fair and ethically sound language models. To this end, our work answers the following research questions:

\begin{enumerate}[label={}, leftmargin=0pt, itemsep=0pt]
    \item \textbf{RQ1:} Compared to traditional methods, does our proposed methodology attain better performance metrics with respect to debiasing SOTA model embeddings?
    \item \textbf{RQ2:} How does our proposed method interact with the varying parameters (size, complexity, embedding dimension) of embeddings obtained from different language models ? 
    \item \textbf{RQ3:} Can we validate that the debiased embeddings, which are a result of our proposed method, are on par with off-the-shelf embeddings on downstream tasks? 
    \item \textbf{RQ4:} How does the type of bias (gender, race, religion) affect the effectiveness of the debiasing process?
\end{enumerate}

To answer the above questions, we make the following contributions through this research:

\vspace{-1mm}
 \begin{tcolorbox}
 [colback=gray!5!white,colframe=gray!75!black,title=\textsc{Our Contributions}]
 \vspace{-2mm}
 \begin{itemize}
 [leftmargin=1mm]
 \setlength\itemsep{0em}
 \begin{spacing}{0.85}
 \vspace{1mm}
 
    \item {\footnotesize 
     {\fontfamily{phv}\fontsize{8}{9}\selectfont
    We provide, to the best of our knowledge, the first comprehensive study of how various debiasing methods work on SoTA large language model word embeddings. }
     } 
     
     \item {\footnotesize 
     {\fontfamily{phv}\fontsize{8}{9}\selectfont
    We present a novel methodology, \textit{DeepSoftDebias}, for debiasing LLM word embeddings, which beats SoTA debiasing methods across multiple bias formats including gender, race, and religion. 
    }
     }

     \item {\footnotesize 
     {\fontfamily{phv}\fontsize{8}{9}\selectfont
     We perform an exhaustive quantitative analysis, establishing SoTA baselines and leveraging multiple evaluation metrics to provide a comparison against accessible SoTA baselines.}
     }    
     
\vspace{-5mm}    
\end{spacing}    
 \end{itemize}
\end{tcolorbox}

\begin{figure*}[h]
  \centering
  \includegraphics[width=1\linewidth]{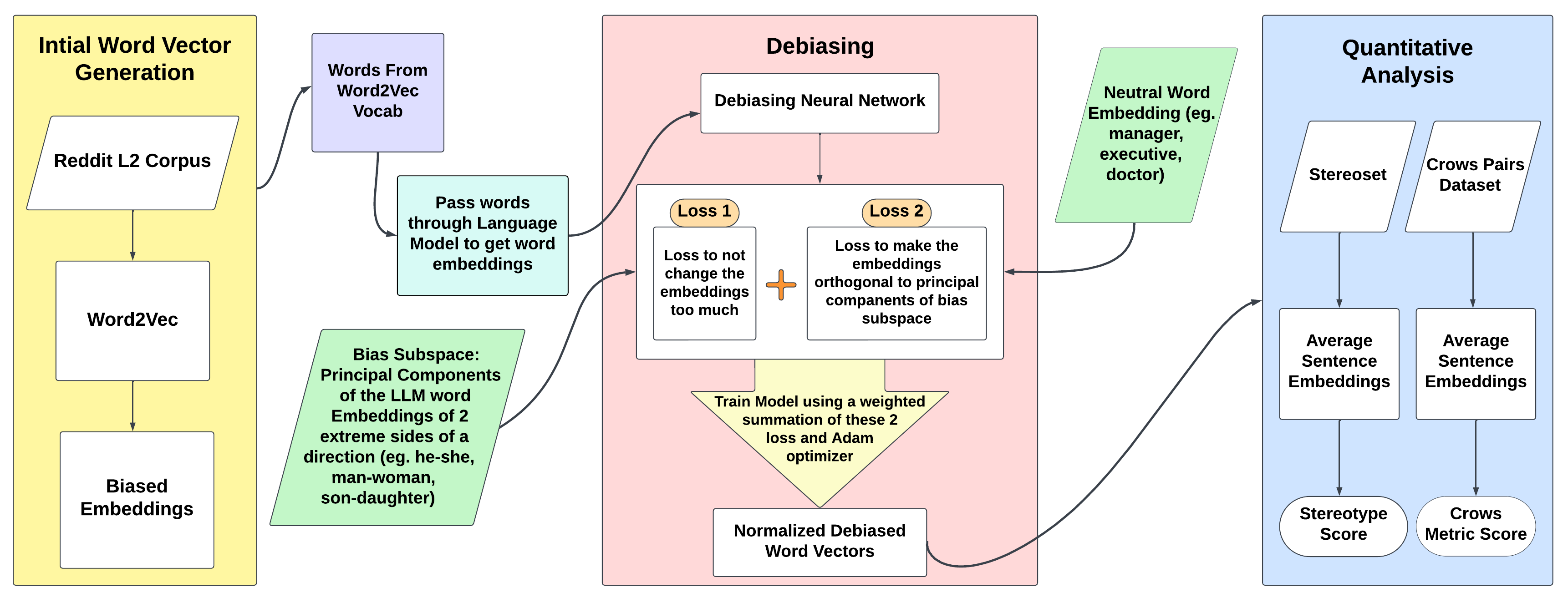} 
  \caption{A step-by-step visualization of the pipeline for \textit{DeepSoftDebias}. Our pipeline has 3 major components, Initial Word Vector Generation, Debiasing, and Quantitative Analysis. The Debiasing stage leverages the \textit{DeepSoftDebias} network.}
  \label{fig:main}
\end{figure*}

We illustrate our pipeline in Fig. \ref{fig:main}. We find that \textit{DeepSoftDebias} not only outperforms the state-of-the-art methods at reducing bias on most of the bias types but also does so while preserving the full information of the original embedding (which is an additional improvement on previous methods). Notably, our proposed methodology is also effective for debiasing embeddings from state space models, namely Mamba \cite{gu2024mambalineartimesequencemodeling}. Further, we find that model performance on challenging downstream tasks like the ones present in the GLUE Benchmark \cite{wang-etal-2018-glue} remains largely unaffected when we test using our debiased embeddings. We also make all of our code available at \url{https://github.com/aishik-rakshit/DeepSoftDebias}

\section{Related Work}

\paragraph{INLP}
Iterative Null-space Projection (INLP) \cite{ravfogel2020null} is a post-hoc debiasing method that operates at the representation level. The INLP methodology debiases representations by iteratively projecting them into a linear classifier’s null space. 
This approach is beneficial in scenarios where an attempt to make a model fairer towards some group results in increased unfairness towards another group. 

\paragraph{Self-Debias}
Self-Debiasing \cite{schick2021selfdiagnosis} is a novel approach to mitigating bias in language models. The methodology is based on the concept of self-diagnosis. In this approach, pretrained language models recognize their undesirable biases and the toxicity of the content they produce. Based on this self-diagnosis, a decoding algorithm is proposed that reduces the probability of a language model producing problematic text. This approach, referred to as self-debiasing, does not rely on manually curated word lists, nor does it require any training data or changes to the model's parameters. While it does not completely eliminate the issue of language models generating biased text, it is an important step in this direction. 

\paragraph{Sentence Debias}
SentenceDebias \cite{liang2020debiasing} is a debiasing methodology that operates at the sentence level. It is a projection-based method that identifies a linear subspace associated with a specific bias. The sentence representations are projected onto this bias subspace, and the projection is subtracted from the original representations. This process effectively debiases the sentence representations. 
,offering a comprehensive comparison between models that adjust weights for debiasing and those employing test-time surgical interventions. 

\paragraph{Counterfactual Data Augumentation}
Counterfactual Data Augmentation (CDA) \cite{yadav2023cda} is a data-based debiasing strategy often used to mitigate gender bias. The CDA methodology involves re-balancing a corpus by swapping bias attribute words (e.g., he/she) in a dataset. This technique is part of a broader set of debiasing techniques that also includes Dropout, Self-Debias, SentenceDebias, and Iterative Nullspace Projection. 

\paragraph{FineDeb}
FineDeb \cite{saravanan2023finedeb} is a two-phase debiasing framework for language models. In the first phase, FineDeb debiases the model by modifying the embeddings learned by the language model. This process involves contextual debiasing of these embeddings. In the second phase, the debiased model is fine-tuned on the language modeling objective. 
Though FineDeb emerges as a robust and effective framework for mitigating bias in language models, it is not open sourced at this time. One of our goals with DeepSoftDebias is to offer an at least equally performative alternative that is open sourced.

\section{Data}
This study leverages several datasets to examine and address biases in word embeddings and language models, focusing on the representation and perpetuation of stereotypes within these systems.

\paragraph{L2-Reddit Corpus}\label{l2-reddit}
We utilize the L2-Reddit\footnote{\url{https://github.com/ellarabi/reddit-l2}} \cite{rabinovich2018native} corpus, a collection of Reddit posts and comments by both native and non-native English speakers, featuring approximately 56 million sentences. This dataset serves as our foundation for training word embeddings, aiming to capture the nuanced and inherently biased linguistic patterns present in social media discourse. In our study, we employ the Reddit L2 corpus as the source for our initial Word2Vec \cite{mikolov2013efficient} word embeddings. Subsequently, we leverage the vocabulary derived from these word vectors to obtain the word embeddings from the LLMs. We utilize Word2Vec on the Reddit-L2 corpus to obtain the vocabulary. This vocabulary comprises the words for which we aim to extract embeddings from the LLMs. The primary objective of this approach is to ensure a consistent set of words across all our LLMs. 


\paragraph{StereoSet}
StereoSet \cite{nadeem2020stereoset} stands out as a critical dataset for measuring stereotype bias in language models, containing around 17,000 sentences across demographic dimensions like gender, race, religion, and profession. It introduces the Context Association Tests (CAT) for evaluating model preferences and biases, providing a structured approach to assess and quantify biases in popular models like BERT \cite{devlin2019bert}, GPT-2 \cite{radford2019language}, RoBERTa \cite{liu2019roberta}, and XLNet \cite{yang2020xlnet}. In our work, we use the Stereoset dataset to benchmark our debiasing method. 

\paragraph{CrowS-Pairs}
CrowS-Pairs \cite{nangia2020crowspairs}, designed to assess social biases in masked language models (MLMs), comprises 1,508 examples covering nine bias types, including race, religion, and age. It contrasts sentences related to historically disadvantaged and advantaged groups in the U.S., with annotations from crowd workers highlighting the degree of stereotyping. In our study, we obtain debiased word embeddings for sentences by computing the average sentence vector for both less and more stereotypical or anti-stereotypical directions. We then compare these embeddings against each other to calculate the Crows Metric score.

\section{Methodology}
In this section, we delve into the domain of debiasing word embeddings, presenting both an established and a newly proposed methodology for mitigating biases in word vector representations. 

\subsection{Bias Identification and Data Structure} 
\label{BiasSpace}
To quantitatively assess bias in word embeddings, we measure the projection of word vectors onto a gender-specific axis, defined by the vector difference between the terms `he' and `she.' The magnitude of this projection serves as an indicator of bias. We use a structured vocabulary with its associated vector representations from the Word2Vec model to facilitate the identification of biases. For a comprehensive evaluation, we utilize additional data files that include definitive sets of gender-associated word pairs, analogy templates that list occupational roles often linked with specific genders, and a set of neutral terms used as evaluation targets. These resources are crucial for the systematic identification and rectification of biases in word embeddings. The words used for the BiasSpace are present in Appendix\ref{bias_words}.

\subsection{Soft Debiasing: The Baseline Approach } 
\label{Soft Debiasing}
The initial method as seen in \cite{manzini2019black} leverages a method called soft debiasing. We recap its algorithm in Algorithm \ref{SoftDebias_OG_Algo}. Soft debiasing involves learning a projection of the embedding matrix that preserves the inner product between biased and debiased embeddings while minimizing the projection onto the bias subspace of embeddings mentioned in \ref{BiasSpace}. Given embeddings $W$ and $N$ which are embeddings for the whole vocabulary and the subset of bias-neutral words respectively, and the bias subspace \textit{B} obtained in Section \ref{l2-reddit}, soft debiasing seeks a linear transformation \textit{A} that minimizes the following objective defined in Eq. \eqref{eq0} as follows:
\begin{equation}
\label{eq0}
\resizebox{0.85\hsize}{!}{
$\left\| (AW)^T(AW) - W^TW \right\|_F^2 + \lambda \left\| (AN)^T(AB) \right\|_F^2$}
\end{equation}

 Minimizing the first term preserves the inner product after the linear transformation \textit{A}, and minimizing the second term minimizes the projection onto the bias subspace \textit{B} of embeddings. $\lambda$ is a tunable parameter that balances the two objectives. \textbf{W} here refers to the matrix of word embeddings and \textbf{N} refers to the matrix of the embeddings of the neutral space i.e. words that aren't influenced by any bias.

 \begin{algorithm}
\caption{Transformation Matrix Approach}
\KwIn{Biased word embeddings ($\text{emb}_{\text{biased}}$), Bias Subspace ($\text{BiasSpace}$), Neutral word embeddings ($\text{emb}_{\text{neutral}}$)}
\KwOut{Debiased word embeddings}
\BlankLine
Perform Singular Value Decomposition (SVD) on $\text{emb}_{\text{biased}}$ to obtain singular values ($s$) and left singular vectors ($u$)\;
Precompute $t1 = s \cdot u^T$ and $t2 = u \cdot s$\;
Compute norm1 as $\| t1 \cdot (T^T \cdot T - I) \cdot t2 \|_F$\;
Compute norm2 as $\| \text{emb}_{\text{neutral}}^T \cdot T^T \cdot \text{BiasSpace} \|_F$\;
Total loss is a weighted combination of norm1 and norm2\;
Optimize transformation matrix using SGD\;
Output debiased word embeddings after recomputing using $T$ and normalizing\;
\label{SoftDebias_OG_Algo}
\end{algorithm}

\subsection{\textit{DeepSoftDebias}: Our Proposed Approach}
In the original approach introduced by \citet{bolukbasi2016man}, a transformation matrix is utilized and optimized by an optimizer to enable a direct mapping between input and output embeddings. To enhance performance, we propose \textit{DeepSoftDebias}. In this approach, we replace the transformation matrix with a neural network made up of residual blocks \cite{he2015deepresiduallearningimage}, leveraging its capability to represent a sequence of transformation matrices. This adaptation enables the algorithm to handle more complex functions mapping between input and output embeddings. We use the same loss functions as mentioned in the section \ref{Soft Debiasing}. Furthermore, we transition from stochastic gradient descent (SGD \cite{robbins1951stochastic}) to the Adam \cite{kingma2017adam} optimizer, resulting in enhanced efficiency, speed, and optimization quality. We describe our full algorithm in Algorithm \ref{DeepSoftDebias}. While these modifications were implemented, the fundamental aspects of the method remain unaltered, ensuring minimal alterations in embeddings and preserving orthogonality with the bias space. 

Unlike the baseline, which relies on singular value decomposition (SVD) and incurred information loss, \textit{DeepSoftDebias} preserves the full information of the original matrix. Moreover, unlike the baseline, \textit{DeepSoftDebias} can handle large embedding dimensions of more than 4.5k. We demonstrate the effectiveness of \textit{DeepSoftDebias} on various datasets and tasks, and show that it outperforms the state-of-the-art methods in terms of accuracy and efficiency. The reason for the need for a fixed \textit{BiasSpace} is that we adopt the methodology proposed by \citet{bolukbasi2016man}. for the derivation of the bias subspace. \textit{The Fixed BiasSpace is crucial for mitigating bias in word embeddings by providing a fixed subspace representing the direction of bias}. The neural network is trained to make embeddings orthogonal to this subspace, reducing bias while maintaining other semantic relationships. This orthogonality minimizes the projection of words onto the bias subspace.

\begin{figure}[!h]
  \centering
  \includegraphics[width=1\linewidth]{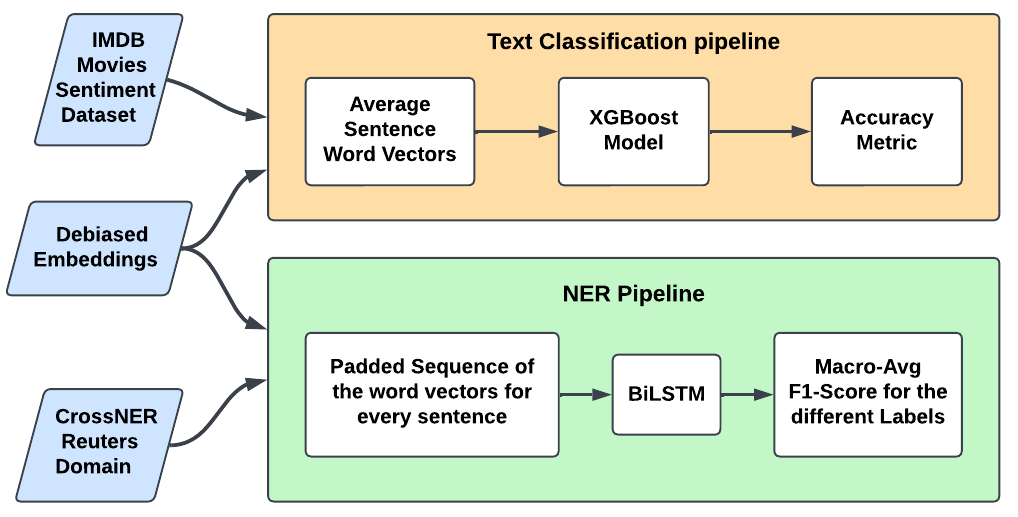} 
  \caption{A step-by-step visualization of our downstream testing process to effectively evaluate \textit{DeepSoftDebias}.}
  \label{fig:downstream}
\end{figure}

The process of creating the \textit{BiasSpace} commences with the identification of word vectors representing opposing concepts, such as ‘he’ versus ‘she’, or ‘man’ versus ‘woman’. For each pair, we compute the mean vector, which encapsulates the shared semantic space. Subsequently, we subtract this mean vector from the original word vectors, yielding vectors that exclusively represent the bias components. These bias vectors are then concatenated to form a matrix, referred to as the bias subspace. This bias subspace plays a pivotal role in the training of our neural network. Specifically, we ensure that the output of the word embeddings, upon being processed through the neural network, is orthogonal to the bias subspace Fig. \ref{fig:downstream} presents a visualization of our approach to downstream testing. Our methodology also extends to multi-dimensional bias representations across race, religion, and other social dimensions. For each group set, we compute a neutral reference point by calculating the mean vector of representative terms. For instance, in racial bias, we derive a neutral vector from terms like "black," "caucasian," and "asian," while for religious bias, we use terms representing different faith traditions such as "judaism," "christianity," and "islam."
We then compute bias-specific axes by subtracting this neutral reference vector from each group's representative vector, yielding vectors that exclusively capture bias components. By taking the vector differences—such as $v_asian$ - $v_neutral$, $v_white$ - $v_neutral$, and $v_black$ - $v_neutral$—we create directional axes that quantify bias in the embedding space. These bias vectors are concatenated to form a comprehensive bias subspace matrix, which plays a pivotal role in the training of our neural network.
Specifically, we ensure that the output of the word embeddings, upon being processed through the neural network, is orthogonal to this bias subspace. This approach allows us to systematically identify, measure, and mitigate bias across multiple social dimensions, moving beyond binary conceptualizations to a more nuanced representation of semantic prejudices. Figure \ref{fig:downstream} presents a visualization of our approach to downstream testing, illustrating how our method can be applied to diverse contextual scenarios.
We also provide a schematic of an example of the transformation Neural Network in Figure \ref{fig:NN_schema}
\begin{algorithm}
\caption{Neural Network Approach}
\KwIn{Biased word embeddings ($\text{emb}_{\text{biased}}$), Bias Subspace ($\text{BiasSpace}$), Neutral word embeddings ($\text{emb}_{\text{neutral}}$)}
\KwOut{Debiased word embeddings}
\BlankLine
Initialize neural network $NN$ with input dimension as embedding dimension and output dimension as embedding dimension\;
Pass $\text{emb}_{\text{biased}}$ through $NN$ to obtain transformed embeddings\;
Compute $T^T$ as the matrix multiplication of the transpose of outputs of $NN$ and the outputs\;
Compute norm1 as $\| (T^T \cdot T - I) \|_F$\;
Compute norm2 as $\| \text{emb}_{\text{neutral}}^T \cdot T^T \cdot \text{BiasSpace} \|_F$\;
Total loss is a weighted combination of norm1 and norm2\;
Optimize $NN$ using an Adam optimizer\;
Output normalized embeddings obtained after passing $\text{emb}_{\text{biased}}$ through $NN$\;
\label{DeepSoftDebias}
\end{algorithm}

\section{Effects of LLM Size and Dependency of Network Size}
The debiasing performance of word embeddings depends on the size of the embeddings and the depth of the debiasing neural network, rather than the number of parameters of the language model. We observe in \ref{fig:nn} Smaller models, such as bge-small \cite{bge_embedding}, DeBERTa-v3-base \cite{he2023debertav3} or DeBERTa-v3-large, GPT2\cite{radford2019language} and Roberta\cite{DBLP:journals/corr/abs-1907-11692} can be debiased effectively by a single-layer neural network. Larger models, such as Llama-2 \cite{touvron2023llama2}, Llama-3 \cite{llama3modelcard}, Alpaca \cite{alpaca} and Yi-6b \cite{yi}, Gemma\cite{gemmateam2024gemmaopenmodelsbased}, Qwen\cite{bai2023qwentechnicalreport} and Mamba\cite{gu2024mambalineartimesequencemodeling} need a more complex debiasing neural network. For embeddings with an embedding length of around 2000, a two-layer neural network is sufficient, while for larger embedding dimensions, a three-layer neural network is required to achieve good debiasing results. In addressing the second research question, we delve into the intricacies of neural network complexity necessary for debiasing embeddings of varying sizes. While our discussion highlights the effectiveness of larger neural networks in mitigating bias within Language Model (LM) embeddings with substantial dimensions, it is imperative to substantiate this observation. We would like to point out that we draw inspiration from the conceptual framework of DeepSoftDebias. Building upon the foundational work by Bolukbasi et al., which employed a transformation matrix for word embedding debiasing, our approach replaces this matrix with a neural network. This neural network can be conceptualized as a series of interconnected matrices. Specifically, when de-biasing larger LMs with embedding dimensions exceeding 4096, we augment the neural network by increasing the number of layers and adjusting layer sizes. This augmentation enables us to model the intricate dependencies inherent in debiasing processes for larger embedding dimensions. Consequently, deeper neural networks emerge as more efficacious tools for addressing bias in such expansive models. Additionally, the debiasing neural network and the optimization algorithm need to be hyperparameter-tuned, such as adjusting the learning rate, to get optimal results. The hyperparameters may vary depending on the model size, the embedding dimension, and the debiasing task. Therefore, a systematic search for the best hyperparameters is necessary to ensure the effectiveness of the debiasing process. 

\section{Results}
In this section, we provide an extensive analysis of our proposed methodology, complete with a comprehensive evaluation against multiple metrics, tasks, and datasets. We provide the results of additional downstream testing and ablation experiments in Appendix~\ref{dtr} and Appendix~\ref{abl}, respectively. We also provide our hypothesis of why there is a variation in bias across LLMs in Appendix~\ref{var}.

\begin{table*}[!htp]\centering
\scriptsize
\begin{tabular}{lrrrrrrrr}\toprule
Bias Type &Model Name &Biased MAC &Soft-Debiased MAC &DSD MAC &Self-Debias MAC &INLP MAC &Sent-Debias MAC \\\midrule
\multirow{6}{*}{Gender} &Bert-Large-Uncased &0.343 &0.734 &0.917 &- &0.720 &0.453 \\
&Roberta-Base &0.025 &0.195 &0.890 &- &0.624 &0.325 \\
&Gemma-7b &0.540 &0.160 &0.908 &- &- &- \\
&Llama 3 8b &0.254 &- &0.939 &- &- &- \\
&SFR Embedding 2\_R &0.388 &- &0.971 &- &- &- \\
&Mamba 2.8b &0.101 &- &0.919 &- &- &- \\
\midrule
\multirow{7}{*}{Race} &Bert-Large-Uncased &0.440 &0.798 &0.997 &- &0.353 &0.450 \\
&Roberta-Base &0.025 &0.205 &0.863 &- &0.422 &0.524 \\
&GPT2-xl &0.497 &0.910 &0.945 &0.378 &0.494 &0.468 \\
&Gemma-7b &0.546 &0.111 &0.951 &- &- &- \\
&Llama 3 8b &0.252 &- &0.937 &- &- &- \\
&SFR Embedding 2\_R &0.440 &- &0.994 &- &- &- \\
&Mamba 2.8b &0.109 &- &0.958 &- &- &- \\
\midrule
\multirow{7}{*}{Religion} &Bert-Large-Uncased &0.455 &0.745 &0.971 &- &0.326 &0.482 \\
&Roberta-Base &0.025 &0.209 &0.936 &- &0.423 &0.425 \\
&GPT2-xl &0.477 &0.927 &0.977 &0.343 &0.549 &0.561 \\
&Gemma-7b &0.372 &- &0.907 &- &- &- \\
&Llama 3 8b &0.264 &- &0.951 &- &- &- \\
&SFR Embedding 2\_R &0.455 &- &0.988 &- &- &- \\
&Mamba 2.8b &0.087 &- &0.945 &- &- &- \\
\bottomrule
\end{tabular}
\caption{Debiasing Results showcasing the MAC scores for the 5 different Debiasing Methods}
\label{tab: Debiasing Results}
\end{table*}

\subsection{Mean Average Cosine Similarity}
Mean Average Cosine Similarity (MAC) \cite{manzini2019black} is a metric used to quantify semantic associations between word classes and attributes. MAC takes word embeddings, targets (representing classes), and attributes as inputs. By computing the mean cosine distance between target words and attribute sets, MAC offers a concise measure of semantic proximity. This metric provides valuable insights into the contextual semantics encoded within word embeddings. Table \ref{tab: Debiasing Results} shows that the word embeddings debiased in the direction of race and gender have comparable increases in their average MAC of \textbf{0.64}, whereas word embeddings debiased in the direction of religion have an increase in MAC of \textbf{0.61}. We see that our debiasing procedure categorically moves MAC scores closer to \textbf{1.0}. This indicates an increase in cosine distance. Further, the associated P-values indicate these changes are statistically significant. This demonstrates that our approach for multiclass debiasing decreases bias in the word embeddings. We provide visual representations of the efficiency of \textit{DeepSoftDebias} at removing gender bias, racial bias, and religion bias in Appendix \ref{MAC_plots}.

In this work, we chose to utilize Mean Average Cosine Similarity (MAC) as our primary metric for assessing bias in word embeddings. This decision is informed by the work of \cite{manzini2019black}, who posit that MAC can be viewed as an extension of the Word Embedding Association Test (WEAT), specifically adapted for a multiclass setting. While WEAT is designed to focus on specific associations between word vectors and predefined concepts (such as gender or race), MAC provides a broader perspective by measuring overall similarity patterns across different groups. This makes MAC less sensitive to specific word choices, thereby revealing biases that might be overlooked by WEAT. In essence, both metrics contribute to a comprehensive understanding of bias in word embeddings. However, the use of MAC is particularly beneficial in our research as it complements the findings of WEAT, providing a more holistic view of bias in the data. This approach allows us to capture a wider range of biases, thereby enhancing the robustness of our analysis.

\subsection{Stereotype Score}
Our research focuses on evaluating and mitigating stereotypical bias in NLI tasks using the Stereoset dataset. This dataset comprises pairs of sentences differing only in the substitution of words related to social groups like gender, race, or religion. The objective is to predict their relationship as same, entailment, or contradiction. We introduce a method aimed at reducing bias in word embeddings, with \textbf{Stereotype Score $SS$} values closer to 50 indicating decreased bias. Table \ref{Stereoset} presents \textit{DeepSoftDebias}'s results alongside existing approaches on the Stereoset dataset. Notably, \textit{DeepSoftDebias} achieves the lowest $SS$ across most social groups, demonstrating its effectiveness in bias reduction. Particularly impressive is \textit{DeepSoftDebias}'s performance in the gender and race categories, where it significantly outperforms existing methods. For instance, with the Mamba 2.8b \cite{jiang2023mistral} model, \textit{DeepSoftDebias} achieves an $SS$ of 50 for gender and with the Llama 3 8b model it acheives a SS of 49.8 for race. We present these scores in Table \ref{Stereoset} and an illustration of these scores in Fig. \ref{fig:stereoset1}. 

\begin{table}[h]
\centering
\resizebox{0.8\columnwidth}{!}{%
\begin{tabular}{lccc}
\toprule
                \multicolumn{4}{c}{Stereotype Score (SS)}        \\ 
\midrule                
\textbf{Stereoset} & \textbf{Gender} & \textbf{Race}  & \textbf{Religion}       \\
\midrule
FineDeb         & \underline {53.27}    & \underline {50.82}    & \textbf{50.39} \\
CDA             & 59.61          & 56.73          & 58.37          \\
INLP            & 57.25          & 57.29          & 60.31          \\
Self-Debias     & 59.34          & 54.30          & 57.26          \\
Sentence Debias & 59.37          & 57.78          & 58.73          \\
\textit{DeepSoftDebias}      & \textbf{50.00} & \textbf{49.8} & \underline {52.56}    \\ 
\bottomrule
\end{tabular}
}
\caption{StereoSet evaluation. Closer to 50 is better for SS. The best performance is highlighted in \textbf{bold} while the next best is \underline{underlined}).}
\label{Stereoset}
\end{table}

\subsection{Crows-Pairs Dataset}
Our study also evaluates social bias in natural language generation tasks using the CrowS Pairs dataset, comprising pairs of sentences differing in their degree of bias. By ranking these sentences according to bias level, we quantify the effectiveness of various methods in reducing bias in word embeddings. 
But as our work is based on word embeddings instead of getting the log-likelihood of the next token from the language model, we compute the average sentence vector for the common parts shared between two sentences.
Next, we compare the similarity of this average sentence vector with the uncommon part (i.e., the modified tokens) using word embeddings. By doing so, we capture the semantic differences between stereotypical and non-stereotypical components within the sentence pairs. The rest of the metric remains the same.

Table \ref{Crows Pairs} presents \textit{DeepSoftDebias}'s results alongside existing approaches on the CrowS Pairs dataset. Notably, \textit{DeepSoftDebias} achieves scores closest to 50 across \textbf{all social groups}, indicating a significant reduction in social bias.
The metric used here is defined in Eq. \eqref{eq1} as follows:
\begin{equation}\label{eq1}
\resizebox{0.89\hsize}{!}{$%
\text{Metric score: } \frac{{(\text{{stereo\_score}} + \text{{antistereo\_score}}) \times 100}}{N}
$%
}%
\end{equation}

where \textbf{Crows Pair Stereotype Score (CSS)} is the number of stereotypical samples that agree with their label direction and \textbf{Crows Pairs Anti-stereotype Score (CAS)} is the number of anti-stereotypical samples that agree with their label direction. Label direction refers to the label given the pair of sentences whether they are stereotypical or anti-stereotypical. In our evaluation we get the average sentence vector of the context and the more and less (anti-)stereotypical sentence. We then see whether the context vector is closer to the more (anti-)stereotypical  sentence or the less (anti-)stereotypical sentence. If it is closer to the more (anti-)stereotypical sentence2901, then we state that it agrees with the (anti-)stereotype, i.e., the label direction.
Particularly noteworthy is \emph{DeepSoftDebias}'s superior performance in the gender and religion categories. For instance, with the mamba-1.4b model, \textit{DeepSoftDebias} achieves a score of \textbf{50.38} for gender and \textbf{50.48} for religion with the mamba-2.8b model. Similarly, using the bge-base-en v1.5 model, \textit{DeepSoftDebias} achieves a score of \textbf{50.19} for debiasing for the bias-type of race. These results underscore the effectiveness of \textit{DeepSoftDebias} in mitigating social bias in word embeddings. We present these scores in \ref{Crows Pairs} and \ref{tab: Annex Debiasing Results} and depict the variation of these scores in Fig. \ref{fig:stereoset2}.

\begin{table}[]
\centering
\resizebox{0.9\columnwidth}{!}{%
\begin{tabular}{lccc}
\toprule
                    \multicolumn{4}{c}{Crows Pairs Metric Score (CMS)}     \\ 
\midrule                       
\textbf{Crows Pairs Dataset} & \textbf{Gender} & \textbf{Race}  & \textbf{Religion}       \\
\midrule
FineDeb             & 54.58          & 65.24          & \underline {44.76}    \\
CDA                 & 56.11          & \underline {56.70}    & 60.00          \\
INLP                & \underline{51.15} & 67.96          & 60.95          \\
Self-Debias         & 52.29          & \underline {56.70}    & 56.19          \\
Sentence Debias     & 52.29          & 62.72          & 63.81          \\
\textit{DeepSoftDebias}          & \textbf{50.38} & \textbf{50.19} & \textbf{50.48} \\ \bottomrule
\end{tabular}
}
\caption{Crows Pairs evaluation. Metric score for every demographic. Closer to 50 is better for the metric (\textbf{best}; \underline{next best}).}
\label{Crows Pairs}
\end{table}


\subsection{Downstream Testing}
For our downstream evaluation, we utilized the GLUE benchmark \cite{wang-etal-2018-glue}. We present the performance differentials between the original word embeddings and their debiased counterparts, processed through various debiasing methods focusing on the three categories of GLUE tasks: single-sentence tasks, sentence-pair tasks, and inference-based tasks.
For the single-sentence task, specifically the Stanford Sentiment Treebank (SST), we report accuracy as the primary metric in Table \ref{tab: SST Main paper}. In the case of sentence-pair tasks, exemplified by the Microsoft Research Paraphrase Corpus (MRPC), we use the F1 score as our performance indicator in Table \ref{tab: Main Paper MRPC}. For the inference-based tasks, which include QNLI (Question-answering Natural Language Inference), WNLI (Winograd Natural Language Inference), RTE (Recognizing Textual Entailment), and MNLI (Multi-Genre Natural Language Inference), we report the average delta of the F1 scores across these four tasks in Table \ref{tab: Inference Type}. For inference-based tasks we see a average gain in performance of 0.017 F1 Score for gender, followed by 0.027 for religion, and a net average performance delta of 0 for race. For sentence pair tasks, we see a performance delta of 0.047 for gender, 0.035 for race and -0.037 F1 score for religion. 
Our results indicate that the DeepSoft Debias method yields performance comparable to that of the original biased word embeddings, with variations in performance ranging from approximately 2\% to 4\% across the different metrics. We hypothesize that this outcome is attributable to the varying degrees of bias present in the datasets, which the models have become less susceptible to following the debiasing process.
We have provided a detailed score $\Delta$s for all the models we have applied out DeepSoftDebias debiasing method on with the GLUE Tasks in Tables \ref{tab: SST Annex}, \ref{tab: STSB Annex}, \ref{tab: MRPC Annex}, \ref{tab: MNLI Annex}, \ref{tab: RTE Annex}, \ref{tab: Annex WNLI} and \ref{tab: Annex QNLI} in the Appendix.
\section{Discussion}
In this section, we summarise the answers to our research questions. 

\paragraph{RQ1}
We find that \textit{\textit{DeepSoftDebias} outperforms state-of-the-art methods, and does so without negatively affecting downstream task performance.} We make this conclusion after exhaustive testing on several models, and datasets and evaluating several metrics. 

\paragraph{RQ2}
We find that \textit{size and complexity do affect the ability of debiasing models.} Specifically, we make the following observations about \textit{DeepSoftDebias}:
\begin{itemize}
    \item A single residual block(RB) neural network can effectively de-bias embeddings with $\text{dim} \leq 1024$.
    \item A two RB neural network can effectively debias embeddings with $\text{dim} \leq 2048$. 
    \item A three RB neural network with an increased layer size can effectively de-bias embeddings with $\text{dim} \leq 4450$.
    
\end{itemize}
With respect to future work, we are curious to investigate scaling patterns to a further extent. A visualization of this is provided in Fig \ref{fig:nn}. We offer a further detailed discussion on hyperparameter tuning the DeepSoftDebias Method in Question 5 of our FAQ section.
2\paragraph{RQ3}
While debiasing techniques in general can affect the downstream performance of models, we test \textit{DeepSoftDebias} on multiple challenging downstream tasks and report that \textit{our proposed approach, to a large extent, does not negatively influence the performance of different downstream tasks. Remarkably, we see an improvement when using our debiased embeddings for some downstream tasks.}

\paragraph{RQ4}
\textls[-10]{We find that while \textit{DeepSoftDebias} is \textit{effective at reducing bias across gender, race, and religion.} We conclude this after testing on multiple embeddings, and multiple datasets and evaluating on multiple performance metrics. As a step for future work, we are curious to investigate whether our proposed approach works towards other forms of bias as well.} We offer a further detailed quantitative discussion of the bias types it works best with, in Question 6 of our FAQ Section.

\section{Conclusion}
\textls[-10]{In this paper, we propose \textit{DeepSoftDebias}, an approach that leverages neural networks to reduce bias in large language model embeddings. We perform an exhaustive series of tests using multiple performance metrics, state-of-the-art datasets, and downstream tasks to ensure that our debiasing technique is robust, efficient, and accurate. In the future, it would be interesting to see how this method translates to multilingual datasets since bias is language and culture-specific. We hope that this research paves the way for future endeavors that look to make LLMs fair, ethical, and bias-free.}

\section{Limitations}
While we do perform exhaustive analysis to test our proposed methodology, our study is monolingual and covers datasets only in English. Consequently, our downstream tasks are also tested only in English. Further, we were unable to conduct test on API-based models at this time. Our testing was also constrained by the limitations of GPU VRAM, which prevented us from extending our testing to larger models such as Llama-65B. These models could not be accommodated within the GPU VRAM, even after applying quantization to 8 bits. Consequently, the largest model that we were able to test was the Gemma-2-9b model.

\section{Ethics Statement}
We understand that bias can be defined in various ways, and it's not necessarily ideal for a language model to treat all users exactly the same without considering demographics. There are situations where certain topics require careful handling to avoid perpetuating harmful stereotypes against marginalized communities. Using specific bias metrics might suggest they encompass all negative social impacts across different groups, but we recognize that existing metrics may not capture all nuances in treatment across demographics. Therefore, any benchmark for bias needs to continually evolve to better understand and address these issues as they affect different communities. 

The definitions of morality and bias are shaped by cultural perspectives, resulting in diverse interpretations among individuals. Consequently, we do not claim that this work provides an objective or exhaustive measure of any of these concepts.

\bibliography{anthology,custom}
\bibliographystyle{acl_natbib}

\newpage
\onecolumn

\section*{Frequently Asked Questions (FAQs)}\label{sec:FAQs}
\begin{enumerate}
    \item \textbf{Is this method effective at removing all kinds of bias?}\\
    We acknowledge that bias has multiple forms that vary by different social factors, language, culture, and various other factors. We evaluated \textit{DeepSoftDebias} on gender bias, racial bias, and religious bias and it has proved effective at reducing all of them. We hope that in the future, this method will prove effective in reducing other kinds of biases as well. 
     \item \textbf{Why isn't GPT analyzed in this paper?}\\
     Given that GPT is an API-based model, we were unable to test it at this time. We hope that one day, this method can be tested even on API-based LLMs. 
    \item \textbf{Is the proposed approach open-sourced?}\\
    Yes, we plan to make all our code available on a GitHub repository. 
    \item \textbf{Why is DeepSoftDebias better than other Debiasing Methods?} \\
    Deep Soft Debiasing (DSD) demonstrates superior performance in mitigating bias in word embeddings compared to other proposed methods such as Iterative Nullspace Projection (INLP), Self-Debias, and Sent-Debias. This superiority is evidenced by the more significant improvement in Mean Average Cosine Similarity (MAC) scores from biased to debiased word embeddings. Furthermore, DSD performs better than or at par with other methods across the three tested bias types: gender, race, and religion, as evaluated on bias detection datasets such as StereoSet and CrowS-Pairs. Notably, DSD maintains the integrity of downstream task performance when utilizing debiased word embeddings, with observed degradation limited to a maximum of 2-3\% compared to biased embeddings. The method's adaptability is a key advantage; it can be readily applied to various models by adjusting hyperparameters such as neural network size, learning rate, and layer dimensions. In contrast, alternative methods often require specific modifications to model architectures, limiting their feasibility across the diverse range of existing models. This flexibility, combined with its robust performance across multiple bias types, positions Deep Soft Debiasing as a more practical and widely applicable approach to addressing bias in natural language processing systems.
    \item \textbf{What is a general hyperparameter tuning starategy for DeepSoftDebais?} \\
    In implementing the Deep Soft Debiasing (DSD) method, four key hyperparameters are crucial for optimizing the debiasing process. The number of residual blocks in the debiasing neural network is adjusted based on the embedding dimension of the target model, with one block sufficing for embedding sizes around 1024, two blocks for sizes around 2048, and three blocks for sizes of 4096 and higher. The learning rate of the optimizer is inversely correlated with the embedding size; larger learning rates (1e-2 to 1e-3) are suitable for models with smaller embedding sizes and consequently smaller debiasing neural networks, while smaller learning rates (1e-4 to 1e-6) are more appropriate for models with larger embedding dimensions. An optional parameter balances two types of losses: projection loss and embedding similarity loss. The default weightage for the neutral space projection loss is 0.2, but this can be increased if the debiasing loss doesn't decrease appreciably for specific models or bias types. While not a direct hyperparameter, the embedding dimension of the model being debiased significantly influences the settings of the other parameters, particularly the number of residual blocks and the learning rate. These interrelated hyperparameters provide the flexibility necessary to adapt the DSD method to various model architectures and bias scenarios, underscoring the importance of careful tuning to achieve optimal debiasing results across different contexts. We also provide a detailed table of the hyperparameters used with each LLM in table \ref{tab: Hyperparameters}
    \item \textbf{Which Bias Categories wiork best with DeepSoftDebias?} \\
     Analysis of Deep Soft Debiasing (DSD) performance across different bias types reveals notable variations in effectiveness. Among the three categories examined, the most substantial improvement in Mean Average Cosine Similarity (MAC) scores was observed for racial bias, with an average increase of 0.63. This was followed by religious bias, showing an average increase of 0.58, and gender bias, with an average increase of 0.56. It is important to note that the absolute average MAC scores after debiasing maintain a distinct hierarchy: racial bias achieves the highest score at 0.95, followed by gender bias at 0.92, and religious bias at 0.89. These results suggest that while DSD demonstrates significant debiasing capabilities across all three categories, its efficacy varies depending on the specific type of bias being addressed. Furthermore, the variation in absolute MAC scores post-debiasing indicates differing levels of residual bias, which may be attributed to the inherent complexities of each bias category or the initial bias severity in the embeddings.

\end{enumerate}

\newpage

\appendix
\section*{Appendix}
\label{sec:appendix}

This section provides supplementary material in the form of additional examples, implementation details, etc. to bolster the reader's understanding of the concepts presented in this work.

\section{Table of words and bias they represent}
\label{bias_words}
\begin{table*}[h]
\centering
\resizebox{0.95\textwidth}{!}{%
\begin{tabular}{llc}\toprule
\textbf{Bias Direction} & & \textbf{Biased Words} \\
\midrule
\multirow{2}{*}{Gender} &Male & \makecell{"manager", "executive", "doctor", "lawyer", "programmer",\\"scientist", "soldier", "supervisor", "rancher", "janitor", "firefighter", "officer"}  \\
&Female & \makecell{"secretary", "nurse", "clerk", "artist", "homemaker", "dancer",\\"singer", "librarian", "maid", "hairdresser", "stylist", "receptionist", "counselor"}  \\
\midrule
\multirow{3}{*}{Race} &Black &"slave", "musician", "runner", "criminal", "homeless" \\
&Caucasian &"manager", "executive", "redneck", "hillbilly", "leader", "farmer" \\
&Asian &"doctor", "engineer", "laborer", "teacher" \\
\midrule
\multirow{3}{*}{Religion} &Jew &"greedy", "cheap", "hairy", "liberal" \\
&Christian &"judgemental", "conservative", "familial" \\
&Muslim &"violent", "terrorist", "dirty", "uneducated" \\
\bottomrule
\end{tabular}
}
\caption{List of Words related to sub-categories in the bias directions explored}
\label{tab:biased_words}
\end{table*}

\section{MAC Scores of \textit{DeepSoftDebias}}
\label{MAC_plots}

\begin{figure*}[h]
  \centering
  \includegraphics[width=1\linewidth]{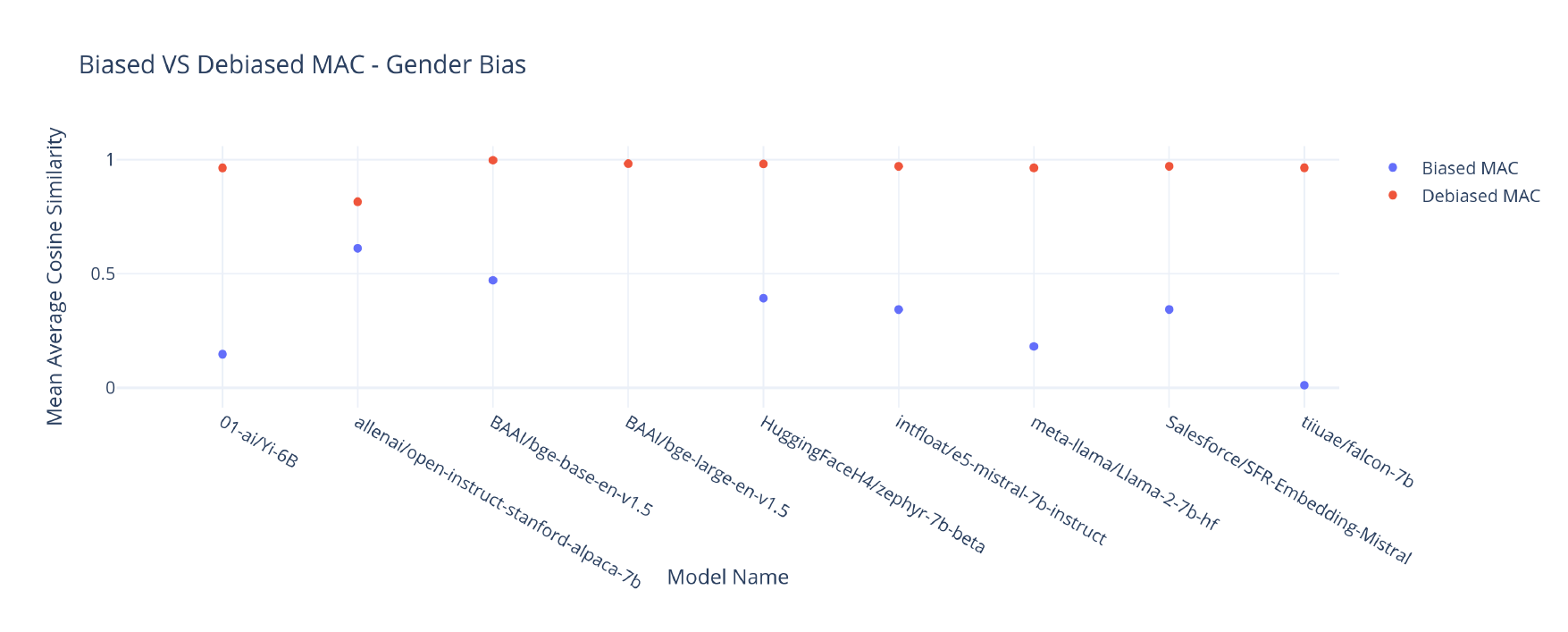} 
  \caption{A visual representation of how \textit{DeepSoftDebias} reduces gender bias in large language model embeddings.}
  \label{fig:MAC_Gender}
\end{figure*}

\begin{figure*}[h]
  \centering
  \includegraphics[width=1\linewidth]{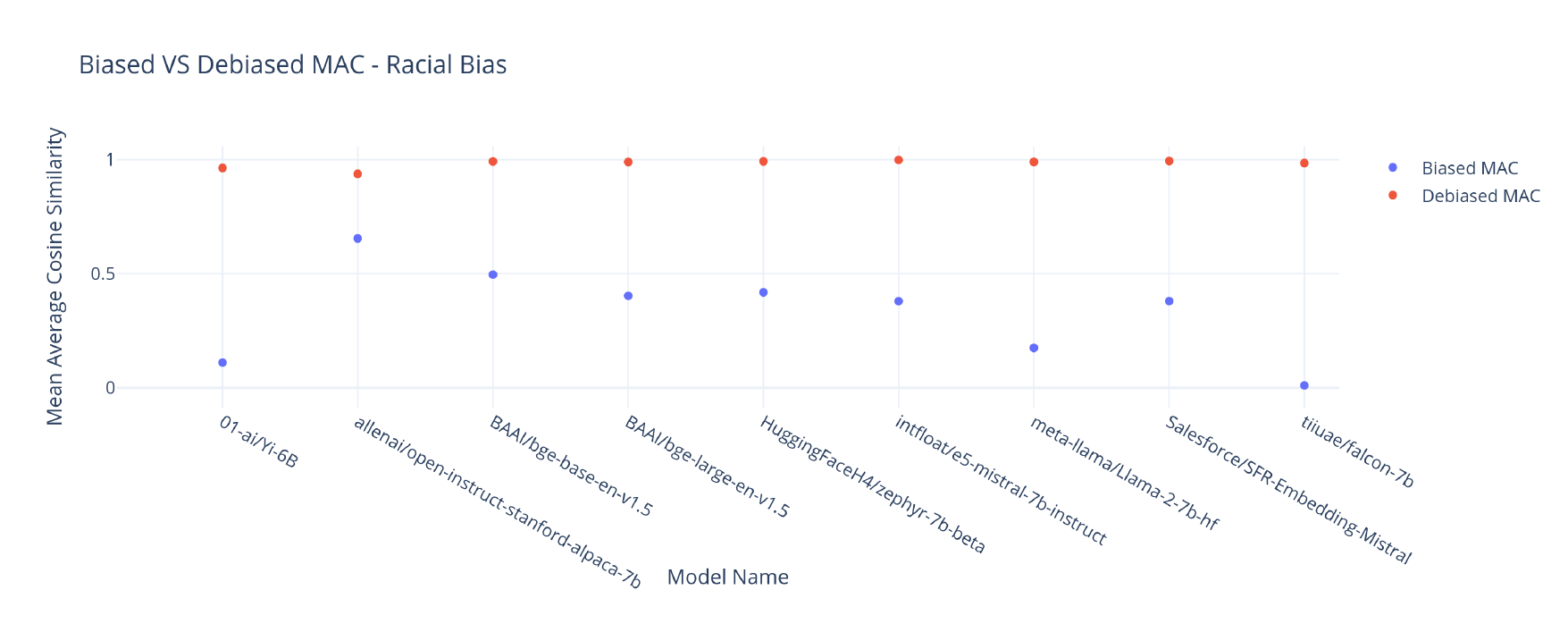} 
  \caption{A visual representation of how \textit{DeepSoftDebias} reduces racial bias in large language model embeddings.}
  \label{fig:MAC_Race}
\end{figure*}

\begin{figure*}[h]
  \centering
  \includegraphics[width=1\linewidth]{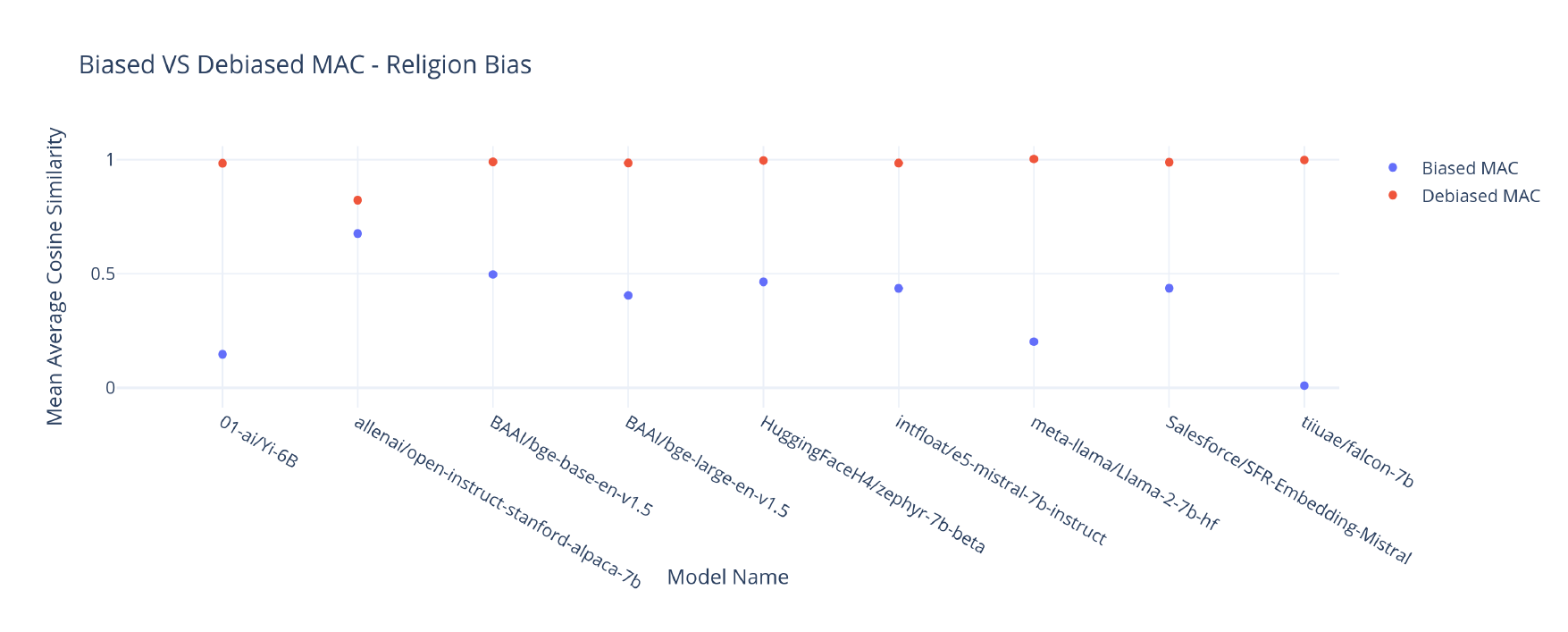} 
  \caption{A visual representation of how \textit{DeepSoftDebias} reduces religion bias in large language model embeddings.}
  \label{fig:MAC_Religion}
\end{figure*}

Figures \ref{fig:MAC_Gender}, \ref{fig:MAC_Race}, and \ref{fig:MAC_Religion} illustrate how \textit{DeepSoftDebias} reduces bias in LLM embeddings. 

\section{Stereoset Scores of \textit{DeepSoftDebias}}

\begin{figure*}[h]
  \centering
  \includegraphics[width=1\linewidth]{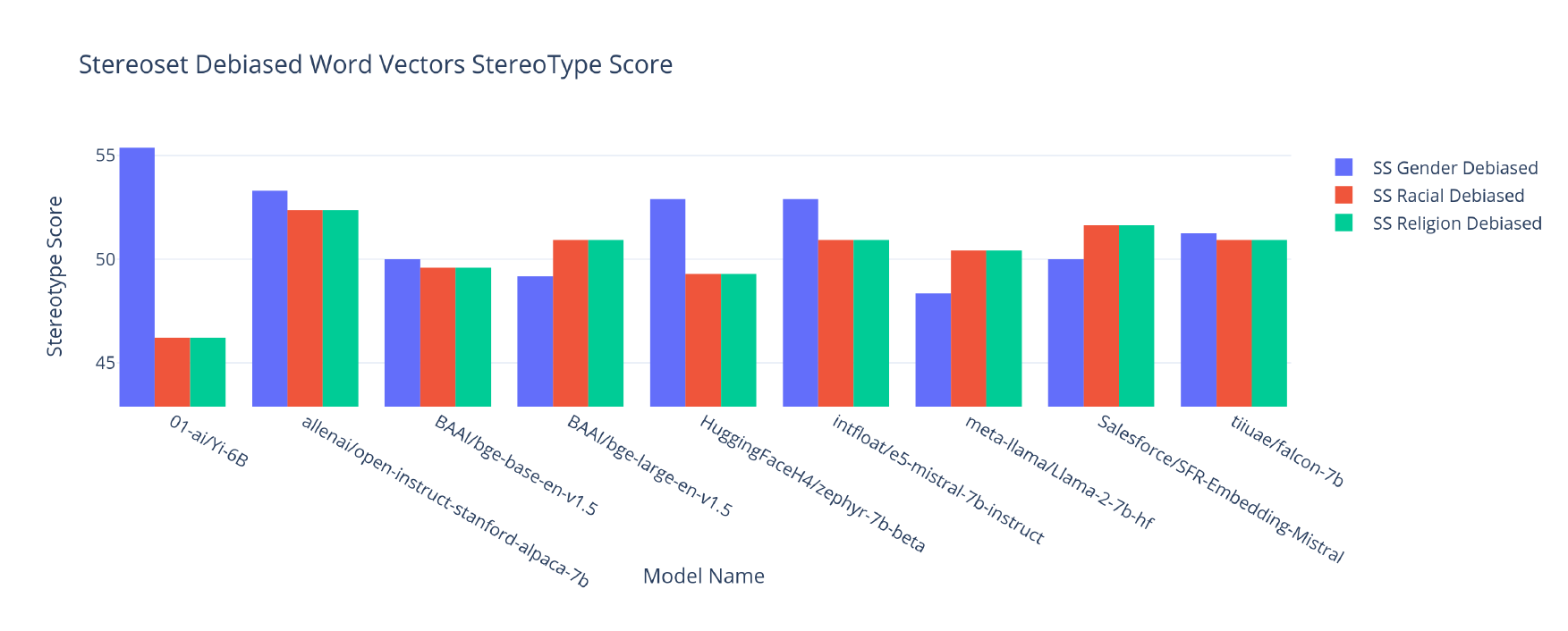} 
  \caption{A visual representation of word vectors debiased using \textit{DeepSoftDebias} and their stereotype scores across gender, race and religion respectively.}
  \label{fig:stereoset1}
\end{figure*}

\begin{figure*}[h]
  \centering
  \includegraphics[width=1\linewidth]{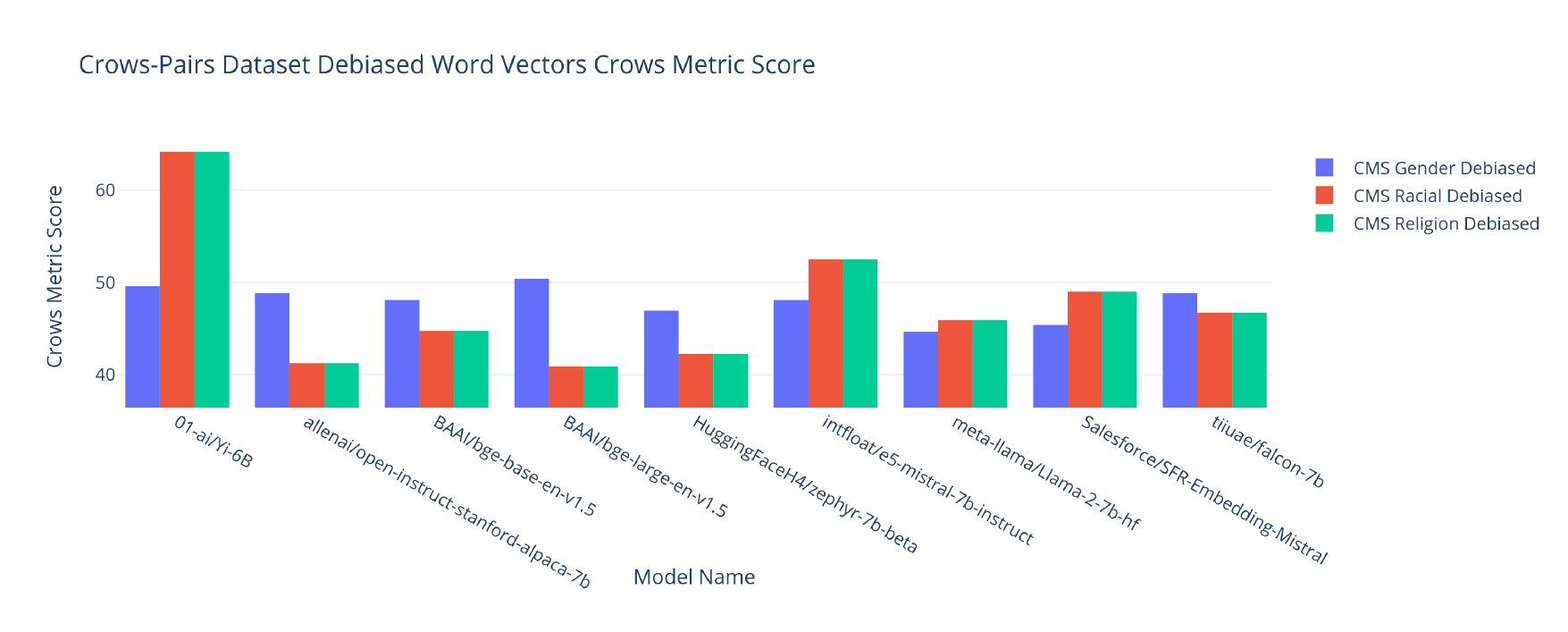} 
  \caption{A visual representation of word vectors debiased using \textit{DeepSoftDebias} and their Crows Metric scores across gender, race and religion respectively.}
  \label{fig:stereoset2}
\end{figure*}

Figures \ref{fig:stereoset1} and \ref{fig:stereoset2} provide an illustration of word vectors debiased using \textit{DeepSoftDebias} and their stereoset scores and Crows Metric scores respectively.

We present the MAC scores, Stereotype Scores, Crows-Pairs Metric Scores in Table \ref{tab: Annex Debiasing Results}
\begin{table}[!htp]
\centering
\scriptsize
\setlength{\tabcolsep}{4pt}
\begin{tabular}{lrrrrrrrrr}\toprule
Bias Type &Model Name &Biased MAC &Soft-Debiased MAC &DSD MAC &Soft-Debiased SS &DSD SS &Soft-Debiased CMS &DSD CMS \\\midrule
\multirow{16}{*}{gender} &gte-Qwen2-7B-instruct &0.469 &- &0.95 &25.21 &45.04 &32.82 &41.22 \\
&bge-base-en-v1.5 &0.447 &0.88 &0.966 &47.93 &43.39 &43.13 &45.04 \\
&bge-large-en-v1.5 &0.408 &0.881 &0.921 &49.59 &48.35 &41.98 &50.76 \\
&roberta-base &0.025 &0.179 &0.922 &52.06 &50.41 &53.44 &51.53 \\
&bert-base-uncased &0.243 &0.408 &0.921 &50.41 &52.07 &40.84 &45.8 \\
&bert-large-uncased &0.324 &0.734 &0.917 &52.89 &48.76 &50 &48.85 \\
&gemma-2-2b &0.46 &- &0.894 &25.62 &49.59 &26.72 &37.4 \\
&gemma-2-9b &0.502 &- &0.915 &25.21 &48.76 &30.15 &46.95 \\
&gemma-2b &0.056 &- &0.847 &19.42 &47.11 &28.24 &41.98 \\
&gemma-7b &0.54 &0.16 &0.908 &45.45 &48.35 &45.04 &45.8 \\
&GritLM-7B &0.38 &- &0.905 &24.38 &49.59 &25.95 &39.31 \\
&Meta-Llama-3-8B &0.254 &- &0.939 &19.83 &50 &23.66 &49.24 \\
&gpt2-xl &0.497 &0.91 &0.945 &49.17 &48.35 &46.18 &49.62 \\
&SFR-Embedding-2\_R &0.388 &- &0.906 &26.03 &51.24 &30.53 &40.08 \\
&mamba-1.4b-hf &0.342 &- &0.935 &43.39 &46.28 &40.08 &50.38 \\
&mamba-2.8b-hf &0.101 &- &0.919 &21.9 &50 &18.32 &51.91 \\
\midrule
\multirow{16}{*}{race} &gte-Qwen2-7B-instruct &0.426 &- &0.971 &26.02 &51.23 &16.86 &63.18 \\
&bge-base-en-v1.5 &0.467 &0.903 &0.987 &51.84 &51.02 &60.27 &50.19 \\
&bge-large-en-v1.5 &0.424 &0.938 &0.99 &51.33 &51.43 &39.34 &50.78 \\
&roberta-base &0.025 &0.205 &0.863 &48.77 &47.44 &30.62 &47.04 \\
&bert-base-uncased &0.245 &0.448 &0.974 &52.25 &51.43 &29.07 &46.07 \\
&bert-large-uncased &0.354 &0.798 &0.997 &49.28 &49.69 &61.43 &44.96 \\
&gemma-2-2b &0.44 &- &0.966 &29.3 &47.95 &41.67 &56.2 \\
&gemma-2b &0.046 &- &0.838 &21.31 &50.31 &12.98 &51.16 \\
&gemma-7b &0.546 &0.111 &0.951 &47.75 &46.72 &53.1 &43.22 \\
&GritLM-7B &0.417 &- &0.971 &21.41 &51.23 &15.5 &46.12 \\
&Meta-Llama-3-8B &0.252 &- &0.937 &22.23 &49.8 &27.91 &43.99 \\
&gpt2 &0.01 &- &0.822 &20.49 &48.77 &12.79 &38.18 \\
&gpt2-xl &0.477 &0.927 &0.977 &49.59 &51.02 &50.39 &55.62 \\
&SFR-Embedding-2\_R &0.44 &- &0.99 &29.2 &51.74 &32.75 &59.88 \\
&mamba-1.4b-hf &0.356 &- &0.994 &45.7 &49.39 &56.98 &47.29 \\
&mamba-2.8b-hf &0.109 &- &0.958 &21.11 &51.33 &19.19 &52.52 \\
\midrule
\multirow{16}{*}{religion} &gte-Qwen2-7B-instruct &0.442 &- &0.975 &21.79 &57.69 &19.05 &70.48 \\
&bge-base-en-v1.5 &0.474 &0.897 &0.987 &50 &47.44 &49.52 &61.9 \\
&bge-large-en-v1.5 &0.412 &0.895 &0.94 &53.85 &55.13 &60 &56.19 \\
&roberta-base &0.025 &0.209 &0.936 &51.28 &56.41 &36.19 &34.29 \\
&bert-base-uncased &0.276 &0.477 &0.991 &56.41 &46.15 &32.38 &68.57 \\
&bert-large-uncased &0.312 &0.745 &0.971 &52.56 &48.72 &42.86 &60.95 \\
&gemma-2-2b &0.523 &- &0.957 &20.51 &52.56 &29.52 &44.76 \\
&gemma-2b &0.095 &- &0.933 &24.36 &55.13 &12.38 &58.1 \\
&gemma-7b &0.372 &- &0.907 &43.59 &52.56 &40.95 &55.48 \\
&GritLM-7B &0.485 &- &0.963 &17.95 &56.41 &25.71 &55.24 \\
&Meta-Llama-3-8B &0.264 &- &0.951 &20.51 &58.97 &20 &52.43 \\
&gpt2 &0.018 &- &0.953 &12.82 &52.56 &22.86 &63.81 \\
&SFR-Embedding-2\_R &0.455 &- &0.959 &34.62 &55.13 &41.9 &54.29 \\
&mamba-1.4b-hf &0.267 &- &0.97 &55.13 &55.13 &41.9 &49.52 \\
&mamba-2.8b-hf &0.087 &- &0.945 &19.23 &46.15 &27.62 &50.48 \\
\bottomrule
\end{tabular}
\caption{Debiasing Results on LLMs: MAC, CMS and SS on all the models we have tested our method on}
\label{tab: Annex Debiasing Results}
\end{table}

\begin{figure*}[h]
  \centering
  \includegraphics[width=1\linewidth]{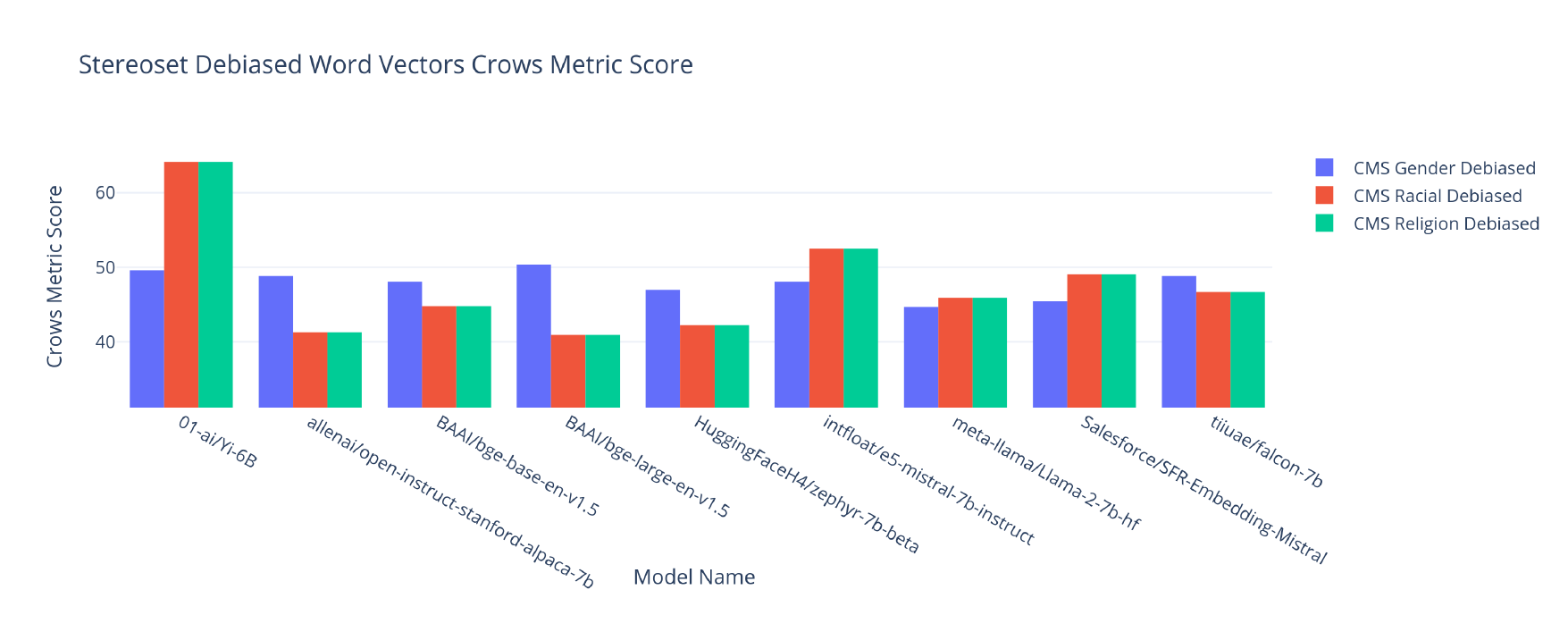} 
  \caption{A visual representation of word vectors debiased using \textit{DeepSoftDebias} and their Crows Metric score across gender, race and religion respectively.}
  \label{fig:stereoset12}
\end{figure*}

\section{Downstream Testing Results}
\label{dtr}


In our research, we primarily focus on the debiasing of word embeddings derived from Language Learning Models (LLMs). We aim to investigate the impact of this debiasing on the performance of these embeddings when subjected to identical training and testing methodologies. Our objective is to quantitatively measure any performance fluctuations (increase or decrease) on the downstream tasks that we test. For this purpose, we trained simple models on top of these word embeddings. For instance, we used an XGBoost model without any hyperparameter tuning for the classification task, and a straightforward bidirectional LSTM for the Named Entity Recognition (NER) task. It is important to note that our goal in presenting our results on these two tasks is not to establish a benchmark for debiased embeddings. Instead, we aim to demonstrate the effect of debiasing on the performance of word embeddings in downstream tasks, as seen in the seminal work of \cite{gonen2019lipstick}. This approach allows us to provide a more comprehensive understanding of the implications and potential benefits of debiasing word embeddings.

\subsection{Text Classification}
In our study, we employ downstream testing to assess the utility of embeddings debiased using \textit{DeepSoftDebias} across two key natural language processing tasks: text classification and named entity recognition (NER). Utilizing the Stanford TreeBank Dataset\cite{socher-etal-2013-recursive} for text classification. Training XGBoost \cite{Chen_2016} classifiers on these vectors, we compare their accuracy on the test set, recognizing accuracy as a straightforward metric for binary classification tasks like sentiment analysis. Notably, our results reveal a slight performance improvement when debiasing in the gender and religion directions inmost cases, whereas a slight decrease in performance is observed in the case of race debiasing in mostr cases. We provide these results in Table:\ref{tab: SST Annex} for Stanford Sentiment Treebank. A visual representation of these results in Fig. \ref{fig:dt1}.

\subsection{Semantic Textual Similarity}

In our research, we evaluate the performance of debiased embeddings for the Semantic Textual Similarity (STS) task using the STS-B dataset. This dataset, a component of the General Language Understanding Evaluation (GLUE) benchmark, is a valuable resource for the STS task. The task aims to quantify the semantic similarity between two sentences, assigning a score from 1 to 5 based on their degree of semantic equivalence. The STS-B dataset, comprising examples from diverse sources, includes human annotations for sentence pair similarity, contributing significantly to the broader field of natural language understanding by facilitating the measurement of meaning equivalence across sentences. To utilize the embeddings for the task, we train a dual-head neural network. We perform cosine similarity after passing the average sentence vector of the two sentences through the network, followed by a Fully Connected layer to obtain the actual score. The performance of our approach is evaluated using Pearson's correlation and Spearman's correlation as metrics. This methodology allows us to develop and evaluate models' ability to understand nuanced semantic relationships in text effectively. We provide our results in this task in Table:\ref{tab: STSB Annex}

\begin{table}[!htp]\centering
\scriptsize
\begin{tabular}{lrrrrrrr}\toprule
Bias Type &Model Name &$\Delta$ Soft-Debias &$\Delta$ Self-Debias &$\Delta$ INLP &$\Delta$ Sent-Debias &$\Delta$ DSD \\\midrule
\multirow{7}{*}{Gender} &Bert-Large-Uncased &1.17 &- &0.453 &0.067 &-0.831 \\
&Roberta-Base &2.75 &- &0.2 &0.252 &0.527 \\
&GPT2 &-7.958 &-3.133 &-0.616 &-0.831 &-6.554 \\
&Gemma-7b &-11.352 &- &- &- &-2.633 \\
&LLama 2 7b &-1.229 &- &- &- &0.059 \\
&SFR Embedding 2\_R &0.995 &- &- &- &0.059 \\
&Mamba 2.8b &-4.623 &- &- &- &-0.995 \\
\midrule
\multirow{7}{*}{Race} &Bert-Large-Uncased &0.995 &- &0.586 &-0.03 &-1.01 \\
&Roberta-Base &-0.293 &- &-0.171 &0.104 &0.761 \\
&GPT2 &-5.5 &-2.992 &-0.23 &-1.01 &-4.389 \\
&Gemma-7b &-10.474 &- &- &- &-1.697 \\
&LLama 2 7b &0.995 &- &- &- &0.41 \\
&SFR Embedding 2\_R &0.936 &- &- &- &1.112 \\
&Mamba 2.8b &-5.734 &- &- &- &0.059 \\
\midrule
\multirow{7}{*}{Religion} &Bert-Large-Uncased &1.053 &- &0.483 &-0.111 &-1.514 \\
&Roberta-Base &1.638 &- &0.23 &-0.245 &1.989 \\
&GPT2 &0.585 &-3.133 &-0.29 &-1.514 &2.575 \\
&Gemma-7b &-13.517 &- &- &- &-2.867 \\
&LLama 2 7b &-1.112 &- &- &- &0.293 \\
&SFR Embedding 2\_R &1.17 &- &- &- &1.58 \\
&Mamba 2.8b &-4.096 &- &- &- &-1.814 \\
\bottomrule
\end{tabular}
\caption{Downstream Result: $\Delta$ of the Accuracy (out of 100) between downstream testing on the SST Dataset using the biased embedding and the Debiased embeddings using various debiasing methods}
\label{tab: SST Main paper}
\end{table}

\begin{table}[!htp]\centering
\scriptsize
\begin{tabular}{lrrrrrrr}\toprule
Bias Type &Model Name &$\Delta$ Soft-Debias &$\Delta$ Self-Debias &$\Delta$ INLP &$\Delta$ Sent-Debias &$\Delta$ DSD \\\midrule
\multirow{7}{*}{Gender} &Bert-Large-Uncased &-0.024 &- &0.032 &-0.113 &-0.017 \\
&Gemma-7b &0.024 &- &- &- &0.008 \\
&GPT2 &0.005 &0.0003 &-0.005 &0.031 &-0.12 \\
&Llama 3 8b &-0.006 &- &- &- &0.035 \\
&Mamba 2.8b &-0.018 &- &- &- &-0.007 \\
&Roberta-Base &0.045 &- &0.032 &-0.113 &0.045 \\
&SFR Embedding 2\_R &0.009 &- &- &- &0.197 \\
\midrule
\multirow{7}{*}{Race} &Bert-Large-Uncased &0.04 &- &-0.001 &0.007 &-0.016 \\
&Gemma-7b &0.015 &- &- &- &0.001 \\
&GPT2 &0.04 &0.036 &-0.068 &0.062 &0.002 \\
&Llama 3 8b &-0.007 &- &- &- &0.07 \\
&Mamba 2.8b &-0.009 &- &- &- &0.009 \\
&Roberta-Base &0.001 &- &-0.001 &0.007 &0.009 \\
&SFR Embedding 2\_R &-0.192 &- &- &- &-0.219 \\
\midrule
\multirow{6}{*}{Religion} &Bert-Large-Uncased &-0.012 &- &-0.052 &-0.04 &0.024 \\
&Gemma-7b &-0.053 &- &- &- &-0.06 \\
&Llama 3 8b &0.002 &- &- &- &0.009 \\
&Mamba 2.8b &0.012 &- &- &- &0.048 \\
&Roberta-Base &0.011 &- &-0.052 &-0.04 &-0.021 \\
&SFR Embedding 2\_R &-0.006 &- &- &- &0.007 \\
\bottomrule
\end{tabular}
\caption{Downstream testing: Average $\Delta$ of the F1 Score between downstream testing on the the 4 Inference type tasks in GLUE (QNLI,WNLI,RTE,MNLI) using the biased embedding and the Debiased embeddings using various debiasing methods}
\label{tab: Inference Type}
\end{table}

\begin{table}[!htp]\centering
\scriptsize
\begin{tabular}{lrrrrrrr}\toprule
Bias Type &Model Name &$\Delta$ Soft-Debias &$\Delta$ Self-Debias &$\Delta$ INLP &$\Delta$ Sent-Debias &$\Delta$ DSD \\\midrule
\multirow{7}{*}{Gender} &Bert-Large-Uncased &-0.162 & &-0.011 &0.002 &-0.146 \\
&Gemma-7b &0.005 & & & &-0.007 \\
&GPT2 &0.166 &0 &-0.005 &-0.004 &-0.005 \\
&Llama 3 8b &-0.007 & & & &-0.16 \\
&Mamba 2.8b &0.078 & & & &0 \\
&Roberta-Base &0.014 & &0.014 &-0.001 &0.152 \\
&SFR Embedding 2\_R &0.146 & & & &0.144 \\
\midrule
\multirow{7}{*}{Race} &Bert-Large-Uncased &0.154 & &-0.003 &0.014 &-0.007 \\
&Gemma-7b &0.238 & & & &0.245 \\
&GPT2 &-0.018 &-0.002 &-0.002 &0.005 &-0.001 \\
&Llama 3 8b &0.153 & & & &-0.242 \\
&Mamba 2.8b &-0.164 & & & &-0.005 \\
&Roberta-Base &-0.014 & &0 &-0.01 &-0.021 \\
&SFR Embedding 2\_R &0 & & & &0.813 \\
\midrule
\multirow{6}{*}{Religion} &Bert-Large-Uncased &-0.175 & &-0.008 &0.005 &-0.42 \\
&Gemma-7b &-0.14 & & & &0.015 \\
&Llama 3 8b &0.018 & & & &0.003 \\
&Mamba 2.8b &0 & & & &-0.006 \\
&Roberta-Base &-0.159 & &0 &-0.01 &-0.006 \\
&SFR Embedding 2\_R &0.241 & & & &0.235 \\
\bottomrule
\end{tabular}
\caption{Downstream testing: $\Delta$ of the F1 Score between downstream testing on MRPC Dataset using the Biased embedding and the Debiased embeddings using various debiasing methods}
\label{tab: Main Paper MRPC}
\end{table}

\begin{table}[!htp]\centering
\scriptsize
\begin{tabular}{lrrrrrrr}\toprule
Bias Type &Model Name &$\Delta$ Soft-Debias &$\Delta$ DSD &$\Delta$ Self-Debias &$\Delta$ INLP &$\Delta$ Sent-Debias \\\midrule
\multirow{14}{*}{gender} &Alibaba-NLP/gte-Qwen2-7B-instruct &0.351 &1.463 &- &- &- \\
&BAAI/bge-base-en-v1.5 &-0.176 &0.351 &- &- &- \\
&BAAI/bge-large-en-v1.5 &-1.872 &-0.644 &- &- &- \\
&FacebookAI/roberta-base &2.75 &0.527 &- &0.2 &0.252 \\
&google-bert/bert-base-uncased &0.468 &-0.41 &- &0.453 &0.067 \\
&google-bert/bert-large-uncased &1.17 &1.755 &- &- &- \\
&google/gemma-2-2b &-6.963 &-1.697 &- &- &- \\
&google/gemma-2b &-1.17 &-0.117 &- &- &- \\
&google/gemma-7b &-11.352 &-2.633 &- &- &- \\
&GritLM/GritLM-7B &-1.931 &-0.995 &- &- &- \\
&openai-community/gpt2-xl &0.585 &2.575 &-4.514 &- &- \\
&Salesforce/SFR-Embedding-2\_R &0.995 &0.059 &- &- &- \\
&state-spaces/mamba-1.4b-hf &-3.862 &2.633 &- &- &- \\
&state-spaces/mamba-2.8b-hf &-4.623 &-0.995 &- &- &- \\
\midrule
\multirow{15}{*}{race} &Alibaba-NLP/gte-Qwen2-7B-instruct &0.585 &1.287 &- &- &- \\
&BAAI/bge-base-en-v1.5 &-0.234 &0.234 &- &- &- \\
&BAAI/bge-large-en-v1.5 &-0.878 &-1.872 &- &- &- \\
&FacebookAI/roberta-base &-0.293 &0.761 &- &-0.171 &0.104 \\
&google-bert/bert-base-uncased &0.527 &-0.351 &- &0.586 &-0.03 \\
&google-bert/bert-large-uncased &0.995 &2.165 &- &- &- \\
&google/gemma-2-2b &-7.139 &-1.872 &- &- &- \\
&google/gemma-2b &-0.468 &0.468 &- &- &- \\
&google/gemma-7b &-10.474 &-1.697 &- &- &- \\
&GritLM/GritLM-7B &-0.702 &-2.224 &- &- &- \\
&openai-community/gpt2 &-7.958 &-6.554 &-2.992 &- &- \\
&openai-community/gpt2-xl &0.819 &1.463 &-4.417 &- &- \\
&Salesforce/SFR-Embedding-2\_R &0.936 &1.112 &- &- &- \\
&state-spaces/mamba-1.4b-hf &-4.74 &1.346 &- &- &- \\
&state-spaces/mamba-2.8b-hf &-5.734 &0.059 &- &- &- \\
\midrule
\multirow{14}{*}{religion} &Alibaba-NLP/gte-Qwen2-7B-instruct &-0.234 &1.58 &- &- &- \\
&BAAI/bge-base-en-v1.5 &-1.287 &0.527 &- &- &- \\
&BAAI/bge-large-en-v1.5 &-2.282 &-2.048 &- &- &- \\
&FacebookAI/roberta-base &1.638 &1.989 &- &0.23 &-0.245 \\
&google-bert/bert-base-uncased &0.585 &-1.521 &- &0.483 &-0.111 \\
&google-bert/bert-large-uncased &1.053 &2.75 &- &- &- \\
&google/gemma-2-2b &-5.968 &-0.995 &- &- &- \\
&google/gemma-2b &-1.697 &0 &- &- &- \\
&google/gemma-7b &-13.517 &-2.867 &- &- &- \\
&GritLM/GritLM-7B &-2.458 &1.229 &- &- &- \\
&openai-community/gpt2 &-5.5 &-4.389 &-3.133 &- &- \\
&Salesforce/SFR-Embedding-2\_R &1.17 &1.58 &- &- &- \\
&state-spaces/mamba-1.4b-hf &-5.734 &1.989 &- &- &- \\
&state-spaces/mamba-2.8b-hf &-4.096 &-1.814 &- &- &- \\
\bottomrule
\end{tabular}
\caption{Downstream testing: $\Delta$ of the Accuracy Score(out of 100) between downstream testing on SST Dataset using the Biased embedding and the Debiased embeddings using various debiasing methods}
\label{tab: SST Annex}
\end{table}

\begin{table}[!htp]\centering
\scriptsize
\begin{tabular}{lrrrrrrr}\toprule
Bias Type &Model Name &$\Delta$ Soft-Debias &$\Delta$ DSD &$\Delta$ Self-Debias &$\Delta$ INLP &$\Delta$ SentDebias \\\midrule
\multirow{14}{*}{gender} &Alibaba-NLP/gte-Qwen2-7B-instruct &-0.041 &-0.04 &- &- &- \\
&BAAI/bge-base-en-v1.5 &-0.06 &-0.083 &- &- &- \\
&BAAI/bge-large-en-v1.5 &0.048 &-0.033 &- &- &- \\
&FacebookAI/roberta-base &0.05 &-0.029 &- &0.029 &-0.024 \\
&google-bert/bert-base-uncased &-0.055 &0.024 &- &-0.014 &-0.024 \\
&google-bert/bert-large-uncased &0.206 &0.157 &- &- &- \\
&google/gemma-2-2b &-0.185 &-0.149 &- &- &- \\
&google/gemma-2b &-0.075 &0.071 &- &- &- \\
&google/gemma-7b &0.008 &0.116 &- &- &- \\
&GritLM/GritLM-7B &-0.069 &-0.092 &- &- &- \\
&openai-community/gpt2-xl &-0.07 &0.099 &0.056 &0.068 &0.04 \\
&Salesforce/SFR-Embedding-2\_R &-0.005 &-0.074 &- &- &- \\
&state-spaces/mamba-1.4b-hf &-0.027 &0.021 &- &- &- \\
&state-spaces/mamba-2.8b-hf &0.051 &-0.011 &- &- &- \\
\midrule
\multirow{15}{*}{race} &Alibaba-NLP/gte-Qwen2-7B-instruct &0.016 &0.212 &- &- &- \\
&BAAI/bge-base-en-v1.5 &0.068 &0.197 &- &- &- \\
&BAAI/bge-large-en-v1.5 &0.04 &0.173 &- &- &- \\
&FacebookAI/roberta-base &-0.015 &0.035 &- &0.011 &-0.022 \\
&google-bert/bert-base-uncased &0.022 &0.197 &- &-0.044 &0.051 \\
&google-bert/bert-large-uncased &-0.005 &0.018 &- &- &- \\
&google/gemma-2-2b &0.043 &0.034 &- &- &- \\
&google/gemma-2b &-0.064 &-0.042 &- &- &- \\
&google/gemma-7b &0.078 &0.026 &- &- &- \\
&GritLM/GritLM-7B &0.021 &0.143 &- &- &- \\
&openai-community/gpt2 &-0.059 &0.069 &-0.126 &-0.128 &0 \\
&openai-community/gpt2-xl &-0.114 &-0.008 &-0.117 &-0.114 &-0.208 \\
&Salesforce/SFR-Embedding-2\_R &0.076 &0.018 &- &- &- \\
&state-spaces/mamba-1.4b-hf &-0.026 &-0.045 &- &- &- \\
&state-spaces/mamba-2.8b-hf &0.007 &0.08 &- &- &- \\
\midrule
\multirow{14}{*}{religion} &Alibaba-NLP/gte-Qwen2-7B-instruct &0.057 &0.016 &- &- &- \\
&BAAI/bge-base-en-v1.5 &0.015 &0.072 &- &- &- \\
&BAAI/bge-large-en-v1.5 &-0.107 &-0.053 &- &- &- \\
&FacebookAI/roberta-base &-0.074 &-0.065 &- &-0.03 &0.016 \\
&google-bert/bert-base-uncased &-0.101 &-0.118 &- &0.048 &-0.04 \\
&google-bert/bert-large-uncased &0.013 &0.154 &- &- &- \\
&google/gemma-2-2b &-0.091 &-0.029 &- &- &- \\
&google/gemma-2b &0.033 &0.024 &- &- &- \\
&google/gemma-7b &0.003 &-0.008 &- &- &- \\
&GritLM/GritLM-7B &-0.061 &-0.026 &- &- &- \\
&openai-community/gpt2 &-0.095 &-0.034 &0.092 &0.059 &0.052 \\
&Salesforce/SFR-Embedding-2\_R &-0.06 &-0.015 &- &- &- \\
&state-spaces/mamba-1.4b-hf &0.03 &-0.005 &- &- &- \\
&state-spaces/mamba-2.8b-hf &0.07 &0.033 &- &- &- \\
\bottomrule
\end{tabular}
\caption{Downstream testing: $\Delta$ of the Pearson Correlation Score between downstream testing on STS-B Dataset using the Biased embedding and the Debiased embeddings using various debiasing methods}
\label{tab: STSB Annex}
\end{table}

\begin{table}[!htp]\centering
\scriptsize
\begin{tabular}{lrrrrrrr}\toprule
Bias Type &Model Name &$\Delta$ Soft-Debias &$\Delta$ DSD &$\Delta$ Self-Debias &$\Delta$ INLP &$\Delta$ Sent-Debias \\\midrule
\multirow{14}{*}{gender} &Alibaba-NLP/gte-Qwen2-7B-instruct &0.151 &0.154 &- &- &- \\
&BAAI/bge-base-en-v1.5 &0.168 &0.156 &- &- &- \\
&BAAI/bge-large-en-v1.5 &0.156 &0.16 &- &- &- \\
&FacebookAI/roberta-base &0.014 &0.152 &- &-0.011 &0.002 \\
&google-bert/bert-base-uncased &-0.151 &-0.001 &- &-0.011 &0.002 \\
&google-bert/bert-large-uncased &0.154 &-0.007 &- &- &- \\
&google/gemma-2-2b &-0.166 &-0.157 &- &- &- \\
&google/gemma-2b &0.003 &-0.145 &- &- &- \\
&google/gemma-7b &0.005 &-0.007 &- &- &- \\
&GritLM/GritLM-7B &-0.155 &0.007 &- &- &- \\
&openai-community/gpt2-xl &0.163 &0.166 &0.005 &0.001 &-0.006 \\
&Salesforce/SFR-Embedding-2\_R &0.146 &0.144 &- &- &- \\
&state-spaces/mamba-1.4b-hf &0.157 &0.005 &- &- &- \\
&state-spaces/mamba-2.8b-hf &0.077 &0.078 &- &- &- \\
\midrule
\multirow{15}{*}{race} &Alibaba-NLP/gte-Qwen2-7B-instruct &0.003 &-0.139 &- &- &- \\
&BAAI/bge-base-en-v1.5 &0.156 &0.167 &- &- &- \\
&BAAI/bge-large-en-v1.5 &-0.001 &0.158 &- &- &- \\
&FacebookAI/roberta-base &-0.014 &-0.021 &- &-0.008 &0.01 \\
&google-bert/bert-base-uncased &0.011 &0.015 &- &-0.003 &0.014 \\
&google-bert/bert-large-uncased &-0.175 &-0.42 &- &- &- \\
&google/gemma-2-2b &0.171 &-0.239 &- &- &- \\
&google/gemma-2b &0.174 &0.161 &- &- &- \\
&google/gemma-7b &0.238 &0.245 &- &- &- \\
&GritLM/GritLM-7B &-0.006 &-0.169 &- &- &- \\
&openai-community/gpt2 &-0.133 &0.012 &-0.002 &-0.002 &0.005 \\
&openai-community/gpt2-xl &-0.417 &-0.018 &0.005 &0.001 &-0.006 \\
&Salesforce/SFR-Embedding-2\_R &0 &0.813 &- &- &- \\
&state-spaces/mamba-1.4b-hf &0.404 &0.404 &- &- &- \\
&state-spaces/mamba-2.8b-hf &-0.001 &-0.164 &- &- &- \\
\midrule
\multirow{14}{*}{religion} &Alibaba-NLP/gte-Qwen2-7B-instruct &0.166 &0.166 &- &- &- \\
&BAAI/bge-base-en-v1.5 &-0.172 &0 &- &- &- \\
&BAAI/bge-large-en-v1.5 &-0.009 &0 &- &- &- \\
&FacebookAI/roberta-base &-0.159 &-0.006 &- &-0.003 &0.014 \\
&google-bert/bert-base-uncased &-0.162 &-0.146 &- &-0.008 &0.005 \\
&google-bert/bert-large-uncased &0.001 &-0.007 &- &- &- \\
&google/gemma-2-2b &0.002 &-0.159 &- &- &- \\
&google/gemma-2b &0.009 &-0.153 &- &- &- \\
&google/gemma-7b &-0.14 &0.015 &- &- &- \\
&GritLM/GritLM-7B &-0.006 &-0.014 &- &- &- \\
&openai-community/gpt2 &0.008 &0.018 &-0.002 &-0.002 &0.005 \\
&Salesforce/SFR-Embedding-2\_R &0.241 &0.235 &- &- &- \\
&state-spaces/mamba-1.4b-hf &0.012 &-0.144 &- &- &- \\
&state-spaces/mamba-2.8b-hf &-0.138 &0 &- &- &- \\
\bottomrule
\end{tabular}
\caption{Downstream testing: $\Delta$ of the F1 Score between downstream testing on MRPC Dataset using the Biased embedding and the Debiased embeddings using various debiasing methods}
\label{tab: MRPC Annex}
\end{table}

\begin{table}[!htp]\centering
\scriptsize
\begin{tabular}{lrrrrrrr}\toprule
Bias Type &Model Name &$\Delta$ Soft-Debais &$\Delta$ DSD &$\Delta$ Self-Debias &$\Delta$ INLP &$\Delta$ Sent-Debias \\\midrule
\multirow{14}{*}{gender} &Alibaba-NLP/gte-Qwen2-7B-instruct &-0.003 &-0.012 & & & \\
&BAAI/bge-base-en-v1.5 &-0.003 &0.005 & & & \\
&BAAI/bge-large-en-v1.5 &-0.001 &-0.008 & & & \\
&FacebookAI/roberta-base &-0.018 &0.001 & &0.001 &0.004 \\
&google-bert/bert-base-uncased &-0.009 &-0.011 & &-0.008 &-0.001 \\
&google-bert/bert-large-uncased &-0.003 &0 & & & \\
&google/gemma-2-2b &-0.002 &0.005 & & & \\
&google/gemma-2b &-0.002 &0.004 & & & \\
&google/gemma-7b &0.004 &-0.015 & & & \\
&GritLM/GritLM-7B &-0.012 &-0.01 & & & \\
&openai-community/gpt2-xl &-0.005 &-0.004 &-0.004 &-0.003 &0.003 \\
&Salesforce/SFR-Embedding-2\_R &0.009 &0.009 & & & \\
&state-spaces/mamba-1.4b-hf &-0.004 &0.012 & & & \\
&state-spaces/mamba-2.8b-hf &0 &0.002 & & & \\
\midrule
\multirow{15}{*}{race} &Alibaba-NLP/gte-Qwen2-7B-instruct &0 &0.001 & & & \\
&BAAI/bge-base-en-v1.5 &-0.008 &0.009 & & & \\
&BAAI/bge-large-en-v1.5 &0.003 &0.012 & & & \\
&FacebookAI/roberta-base &0.001 &-0.003 & &-0.008 &-0.001 \\
&google-bert/bert-base-uncased &-0.001 &0.001 & &-0.008 &-0.001 \\
&google-bert/bert-large-uncased &-0.014 &0.001 & & & \\
&google/gemma-2-2b &-0.002 &0.004 & & & \\
&google/gemma-2b &-0.018 &-0.021 & & & \\
&google/gemma-7b &0.014 &-0.003 & & & \\
&GritLM/GritLM-7B &-0.006 &-0.005 & & & \\
&openai-community/gpt2 &-0.004 &0.005 &0.004 &0.017 &0.004 \\
&openai-community/gpt2-xl &-0.001 &0.005 &-0.004 &-0.003 &0.003 \\
&Salesforce/SFR-Embedding-2\_R &0.013 &0.012 & & & \\
&state-spaces/mamba-1.4b-hf &-0.01 &0.009 & & & \\
&state-spaces/mamba-2.8b-hf &-0.005 &0.001 & & & \\
\midrule
\multirow{14}{*}{religion} &Alibaba-NLP/gte-Qwen2-7B-instruct &-0.011 &0.002 & & & \\
&BAAI/bge-base-en-v1.5 &-0.002 &-0.012 & & & \\
&BAAI/bge-large-en-v1.5 &-0.007 &0 & & & \\
&FacebookAI/roberta-base &0.006 &-0.006 & &-0.008 &-0.001 \\
&google-bert/bert-base-uncased &0 &-0.001 & &-0.005 &0.009 \\
&google-bert/bert-large-uncased &0.001 &0.008 & & & \\
&google/gemma-2-2b &-0.013 &-0.012 & & & \\
&google/gemma-2b &-0.012 &-0.001 & & & \\
&google/gemma-7b &0.003 &-0.005 & & & \\
&GritLM/GritLM-7B &0 &0.008 & & & \\
&openai-community/gpt2 &0.007 &0.011 &0.004 &0.017 &0.004 \\
&Salesforce/SFR-Embedding-2\_R &0.021 &0.011 & & & \\
&state-spaces/mamba-1.4b-hf &-0.011 &0.001 & & & \\
&state-spaces/mamba-2.8b-hf &-0.006 &0.002 & & & \\
\bottomrule
\end{tabular}
\caption{Downstream testing: $\Delta$ of the F1 Score between downstream testing on MNLI Dataset using the Biased embedding and the Debiased embeddings using various debiasing methods}
\label{tab: MNLI Annex}
\end{table}

\begin{table}[!htp]\centering
\scriptsize
\begin{tabular}{lrrrrrrr}\toprule
Bias Type &Model Name &$\Delta$ Soft-Debias &$\Delta$ DSD &$\Delta$ Self-Debias &$\Delta$ INLP &$\Delta$ Sent-Debias \\\midrule
\multirow{14}{*}{gender} &Alibaba-NLP/gte-Qwen2-7B-instruct &0 &0.452 & & & \\
&BAAI/bge-base-en-v1.5 &-0.002 &0.203 & & & \\
&BAAI/bge-large-en-v1.5 &0.005 &0.008 & & & \\
&FacebookAI/roberta-base &0.143 &0.113 & &0.016 &-0.005 \\
&google-bert/bert-base-uncased &-0.099 &-0.101 & &0.033 &0.201 \\
&google-bert/bert-large-uncased &-0.118 &-0.115 & & & \\
&google/gemma-2-2b &0.006 &-0.008 & & & \\
&google/gemma-2b &-0.092 &-0.417 & & & \\
&google/gemma-7b &0.037 &-0.002 & & & \\
&GritLM/GritLM-7B &0.325 &0.411 & & & \\
&openai-community/gpt2-xl &0.075 &-0.333 & & & \\
&Salesforce/SFR-Embedding-2\_R &0 &0.579 & & & \\
&state-spaces/mamba-1.4b-hf &0.012 &0.011 & & & \\
&state-spaces/mamba-2.8b-hf &-0.042 &-0.029 & & & \\
\midrule
\multirow{15}{*}{race} &Alibaba-NLP/gte-Qwen2-7B-instruct &-0.005 &-0.008 & & & \\
&BAAI/bge-base-en-v1.5 &0.043 &0.015 & & & \\
&BAAI/bge-large-en-v1.5 &-0.022 &0.089 & & & \\
&FacebookAI/roberta-base &0.002 &0.011 & &0.011 &0.005 \\
&google-bert/bert-base-uncased &0.03 &0.199 & &0.018 &0.065 \\
&google-bert/bert-large-uncased &0.138 &-0.02 & & & \\
&google/gemma-2-2b &0.008 &0.142 & & & \\
&google/gemma-2b &-0.013 &0 & & & \\
&google/gemma-7b &0.02 &0.006 & & & \\
&GritLM/GritLM-7B &0.01 &0.026 & & & \\
&openai-community/gpt2 &-0.088 &0.008 &0.01 & & \\
&openai-community/gpt2-xl &0.118 &0.019 &-0.046 & & \\
&Salesforce/SFR-Embedding-2\_R &-0.644 &-0.644 & & & \\
&state-spaces/mamba-1.4b-hf &-0.005 &0.107 & & & \\
&state-spaces/mamba-2.8b-hf &-0.013 &-0.011 & & & \\
\midrule
\multirow{14}{*}{religion} &Alibaba-NLP/gte-Qwen2-7B-instruct &-0.009 &0.066 & & & \\
&BAAI/bge-base-en-v1.5 &0.003 &0.001 & & & \\
&BAAI/bge-large-en-v1.5 &0.021 &0.015 & & & \\
&FacebookAI/roberta-base &0.021 &0.005 & &-0.09 &-0.134 \\
&google-bert/bert-base-uncased &0.314 &0.513 & &0.004 &0.005 \\
&google-bert/bert-large-uncased &-0.029 &0.065 & & & \\
&google/gemma-2-2b &0.019 &0.052 & & & \\
&google/gemma-2b &0.012 &0.175 & & & \\
&google/gemma-7b &-0.15 &-0.178 & & & \\
&GritLM/GritLM-7B &-0.009 &-0.018 & & & \\
&openai-community/gpt2 &0.322 &0.338 &-0.014 & & \\
&Salesforce/SFR-Embedding-2\_R &-0.073 &0 & & & \\
&state-spaces/mamba-1.4b-hf &0.007 &-0.014 & & & \\
&state-spaces/mamba-2.8b-hf &0.011 &0.13 & & & \\
\bottomrule
\end{tabular}
\caption{Downstream testing: $\Delta$ of the F1 Score between downstream testing on RTE Dataset using the Biased embedding and the Debiased embeddings using various debiasing methods}
\label{tab: RTE Annex}
\end{table}

\begin{table}[!htp]\centering
\scriptsize
\begin{tabular}{lrrrrrrr}\toprule
Bias Type &Model Name &$\Delta$ Soft-Debias &$\Delta$ DSD &$\Delta$ Self-Debias &$\Delta$ INLP &$\Delta$ Sent-Debias \\\midrule
\multirow{14}{*}{gender} &Alibaba-NLP/gte-Qwen2-7B-instruct &-0.011 &0.011 & & & \\
&BAAI/bge-base-en-v1.5 &-0.011 &0.014 & & & \\
&BAAI/bge-large-en-v1.5 &0.033 &-0.09 & & & \\
&FacebookAI/roberta-base &0.011 &0.021 & &0.078 &-0.339 \\
&google-bert/bert-base-uncased &0 &-0.057 & &0.01 &-0.014 \\
&google-bert/bert-large-uncased &0.05 &0.064 & & & \\
&google/gemma-2-2b &-0.004 &-0.011 & & & \\
&google/gemma-2b &-0.039 &0 & & & \\
&google/gemma-7b &0.03 &0.041 & & & \\
&GritLM/GritLM-7B &-0.011 &0.007 & & & \\
&openai-community/gpt2-xl &-0.056 &-0.024 &0.081 &-0.011 &0.044 \\
&Salesforce/SFR-Embedding-2\_R &0.019 &0.004 & & & \\
&state-spaces/mamba-1.4b-hf &0.014 &0.017 & & & \\
&state-spaces/mamba-2.8b-hf &-0.011 &0.007 & & & \\
\midrule
\multirow{15}{*}{race} &Alibaba-NLP/gte-Qwen2-7B-instruct &0.01 &-0.003 & & & \\
&BAAI/bge-base-en-v1.5 &-0.018 &-0.029 & & & \\
&BAAI/bge-large-en-v1.5 &-0.017 &-0.131 & & & \\
&FacebookAI/roberta-base &0 &0.019 & &-0.005 &0.017 \\
&google-bert/bert-base-uncased &-0.051 &-0.047 & &-0.018 &-0.263 \\
&google-bert/bert-large-uncased &-0.003 &-0.028 & & & \\
&google/gemma-2-2b &0.018 &0.007 & & & \\
&google/gemma-2b &-0.014 &0.01 & & & \\
&google/gemma-7b &0.01 &0 & & & \\
&GritLM/GritLM-7B &-0.018 &0 & & & \\
&openai-community/gpt2 &0.023 &0.015 &0.118 &-0.221 &0.107 \\
&openai-community/gpt2-xl &0.004 &-0.019 &0.014 &0.034 &-0.091 \\
&Salesforce/SFR-Embedding-2\_R &0.056 &-0.024 & & & \\
&state-spaces/mamba-1.4b-hf &0.037 &0.015 & & & \\
&state-spaces/mamba-2.8b-hf &-0.008 &0.037 & & & \\
\midrule
\multirow{14}{*}{religion} &Alibaba-NLP/gte-Qwen2-7B-instruct &-0.035 &-0.028 & & & \\
&BAAI/bge-base-en-v1.5 &-0.014 &0.026 & & & \\
&BAAI/bge-large-en-v1.5 &-0.022 &0.014 & & & \\
&FacebookAI/roberta-base &0.007 &-0.062 & &-0.059 &0.016 \\
&google-bert/bert-base-uncased &0 &0.01 & &-0.018 &-0.029 \\
&google-bert/bert-large-uncased &-0.007 &0 & & & \\
&google/gemma-2-2b &0 &-0.021 & & & \\
&google/gemma-2b &-0.033 &0.006 & & & \\
&google/gemma-7b &-0.011 &0.004 & & & \\
&GritLM/GritLM-7B &-0.037 &-0.007 & & & \\
&openai-community/gpt2 &-0.011 &-0.026 &-0.007 &0.18 &0.003 \\
&Salesforce/SFR-Embedding-2\_R &0.035 &0.009 & & & \\
&state-spaces/mamba-1.4b-hf &-0.018 &0.011 & & & \\
&state-spaces/mamba-2.8b-hf &0.032 &0.011 & & & \\
\bottomrule
\end{tabular}
\caption{Downstream testing: $\Delta$ of the F1 Score between downstream testing on WNLI Dataset using the Biased embedding and the Debiased embeddings using various Debiasing Methods}
\label{tab: Annex WNLI}
\end{table}

\begin{table}[!htp]\centering
\caption{QNLI}\label{tab: }
\scriptsize
\begin{tabular}{lrrrr}\toprule
Bias Type &Model Name &$\Delta$ Soft-Debias &$\Delta$ DSD \\\midrule
\multirow{15}{*}{gender} &Alibaba-NLP/gte-Qwen2-7B-instruct &-0.21 &0.163 \\
&BAAI/bge-base-en-v1.5 &-0.036 &-0.077 \\
&BAAI/bge-large-en-v1.5 &-0.019 &-0.11 \\
&FacebookAI/roberta-base &-0.099 &-0.02 \\
&google-bert/bert-base-uncased &-0.018 &-0.109 \\
&google-bert/bert-large-uncased &0.006 &-0.025 \\
&google/gemma-2-2b &-0.321 &0.021 \\
&google/gemma-2b &-0.256 &0.091 \\
&google/gemma-7b &-0.153 &0.195 \\
&GritLM/GritLM-7B &-0.191 &0.014 \\
&meta-llama/Meta-Llama-3-8B &-0.158 &0.246 \\
&openai-community/gpt2-xl &-0.015 &-0.011 \\
&Salesforce/SFR-Embedding-2\_R &-0.12 &0.17 \\
&state-spaces/mamba-1.4b-hf &-0.32 &0.009 \\
&state-spaces/mamba-2.8b-hf &-0.276 &0.067 \\
\midrule
\multirow{16}{*}{race} &Alibaba-NLP/gte-Qwen2-7B-instruct &-0.199 &0.132 \\
&BAAI/bge-base-en-v1.5 &-0.027 &-0.016 \\
&BAAI/bge-large-en-v1.5 &0.003 &-0.012 \\
&FacebookAI/roberta-base &-0.059 &-0.04 \\
&google-bert/bert-base-uncased &-0.023 &-0.08 \\
&google-bert/bert-large-uncased &-0.05 &-0.106 \\
&google/gemma-2-2b &-0.314 &0.038 \\
&google/gemma-2b &-0.29 &0.077 \\
&google/gemma-7b &-0.141 &0.163 \\
&GritLM/GritLM-7B &-0.173 &0.191 \\
&meta-llama/Meta-Llama-3-8B &-0.146 &0.216 \\
&openai-community/gpt2 &-0.272 &-0.018 \\
&openai-community/gpt2-xl &0.012 &-0.013 \\
&Salesforce/SFR-Embedding-2\_R &-0.16 &0.198 \\
&state-spaces/mamba-1.4b-hf &-0.351 &-0.057 \\
&state-spaces/mamba-2.8b-hf &-0.282 &-0.038 \\
\midrule
\multirow{15}{*}{religion} &Alibaba-NLP/gte-Qwen2-7B-instruct &-0.201 &0.14 \\
&BAAI/bge-base-en-v1.5 &-0.027 &-0.015 \\
&BAAI/bge-large-en-v1.5 &0.023 &0.015 \\
&FacebookAI/roberta-base &-0.059 &-0.023 \\
&google-bert/bert-base-uncased &-0.024 &-0.044 \\
&google-bert/bert-large-uncased &-0.009 &-0.051 \\
&google/gemma-2-2b &-0.314 &0.042 \\
&google/gemma-2b &-0.286 &0.039 \\
&google/gemma-7b &-0.187 &0.166 \\
&GritLM/GritLM-7B &-0.159 &0.19 \\
&meta-llama/Meta-Llama-3-8B &-0.154 &0.177 \\
&openai-community/gpt2 &-0.274 &-0.004 \\
&Salesforce/SFR-Embedding-2\_R &-0.134 &0.23 \\
&state-spaces/mamba-1.4b-hf &-0.3 &-0.016 \\
&state-spaces/mamba-2.8b-hf &-0.245 &0.066 \\
\bottomrule
\end{tabular}
\caption{Downstream testing: $\Delta$ of the F1 Score between downstream testing on WNLI Dataset using the Biased embedding and the Debiased embeddings using Soft-Debiasing and DeepSoftDebias}
\label{tab: Annex QNLI}
\end{table}

\begin{figure*}[h]
  \centering
  \includegraphics[width=1\linewidth]{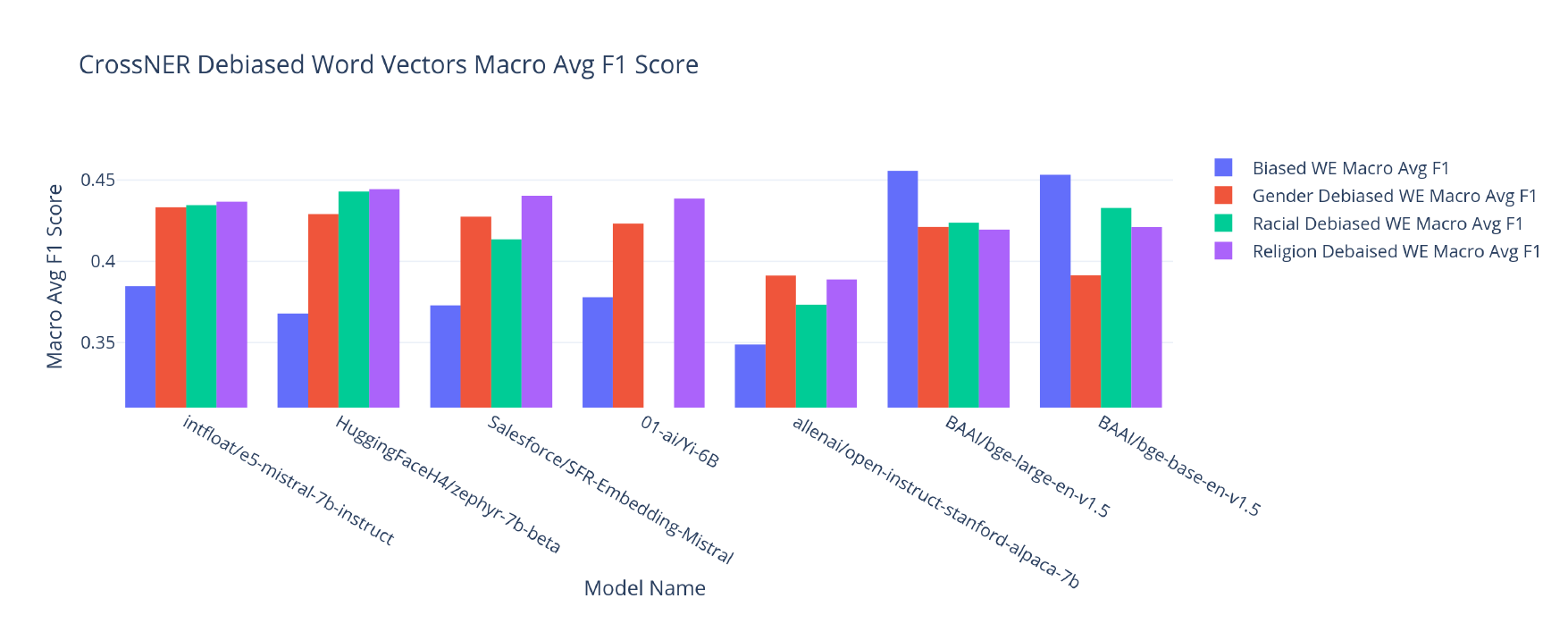} 
  \caption{An illustration of the results of downstream testing on NER. We compare the performance of biased and debaised embeddings in the directions of gender, race, and religion respectively.}
  \label{fig:dt1}
\end{figure*}

\begin{figure*}[h]
  \centering
  \includegraphics[width=1\linewidth]{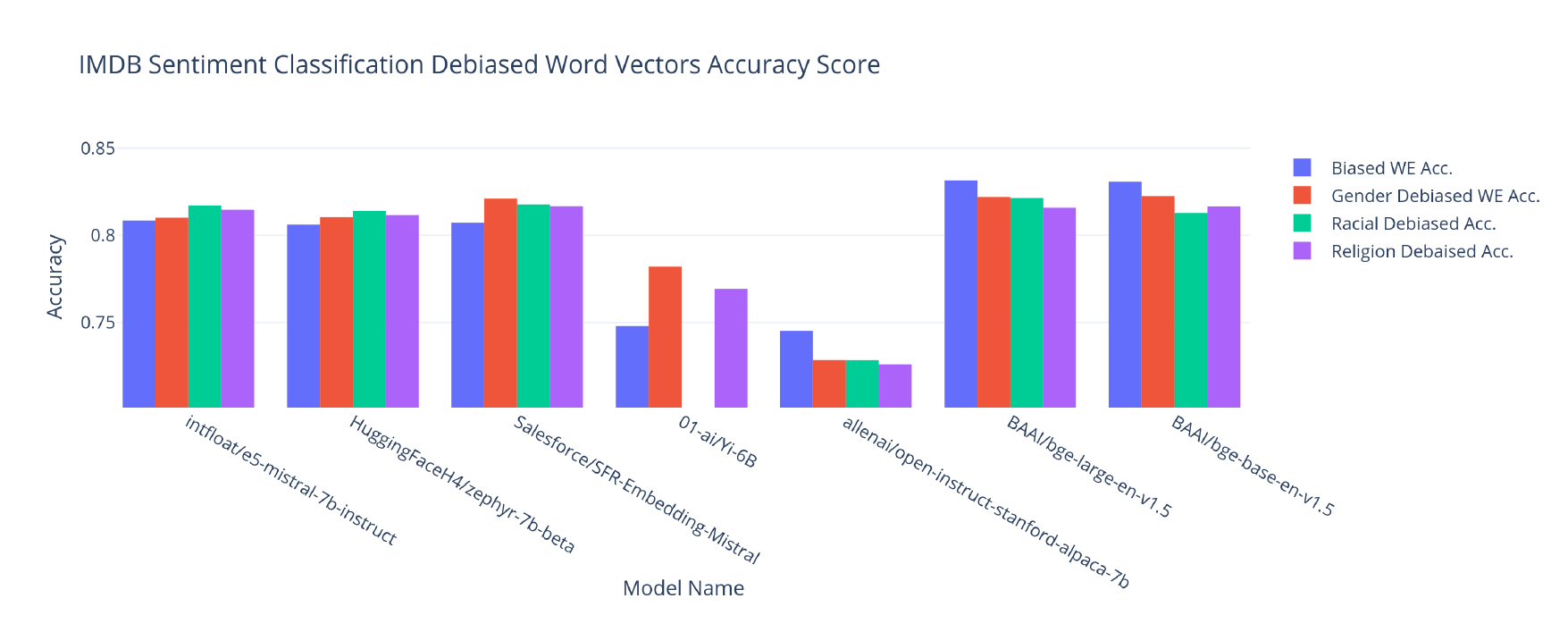} 
  \caption{An illustration of results of downstream testing on sentiment analysis. We compare the performance of biased and debaised embeddings in the directions of gender, race, and religion respectively.}
  \label{fig:dt2}
\end{figure*}

Figures \ref{fig:dt1} and \ref{fig:dt2} present an illustration of the results of various downstream tasks and their performance evaluation.

\begin{figure*}[h]
  \centering
  \includegraphics[width=0.8\linewidth]{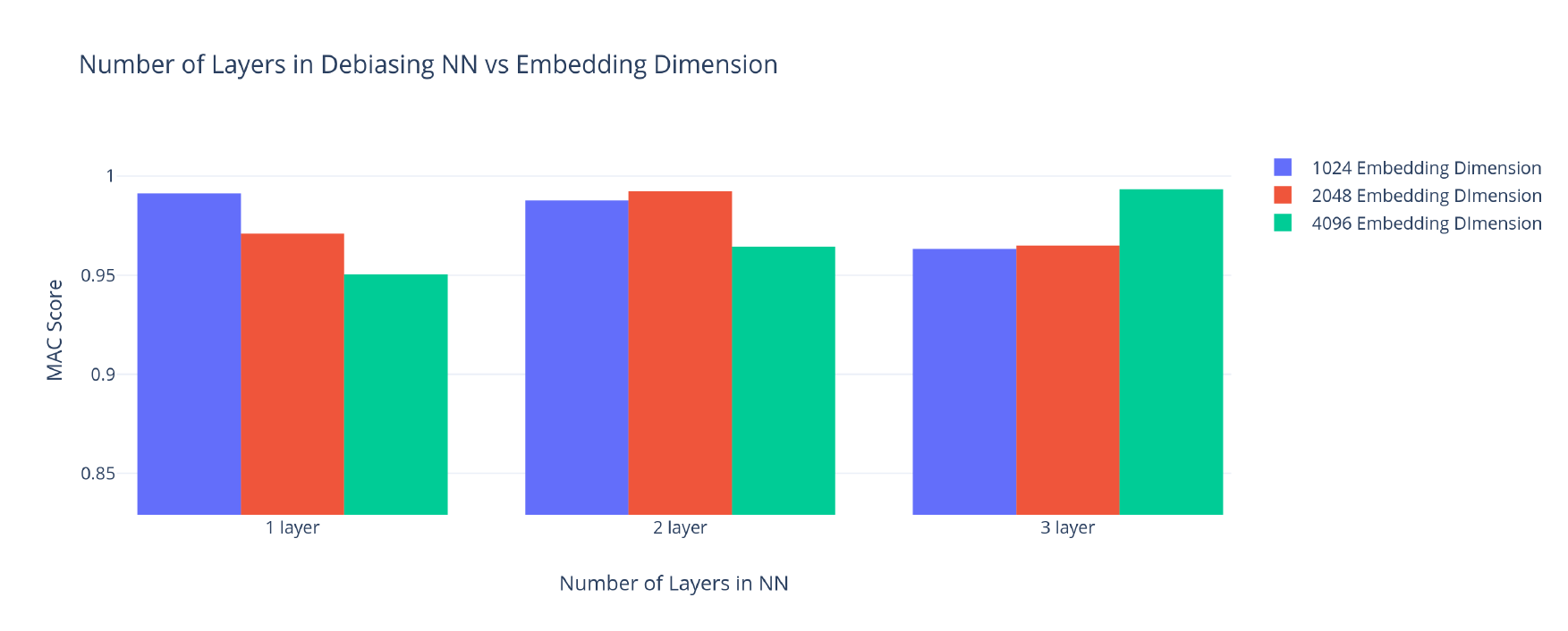} 
  \caption{An illustration analysis of number of layers in debiasing neural network vs. embedding dimension. We can see the varying performance of the 3 different sizes according to the embedding dimension of the LM it is used with.}
  \label{fig:nn}
\end{figure*}

\begin{table*}[h]
\centering
\scriptsize
\setlength{\tabcolsep}{4pt}
\begin{tabular}{lcccccc}
\toprule
\textbf{Model Name} & \textbf{Topic} & \textbf{\makecell{STS-B $\uparrow$ Baseline\\ Debiased PCC}} & \textbf{\makecell{STS-B $\uparrow$ \textit{DeepSoftDebias}\\ Debiased PCC}} & \textbf{\makecell{SST Biased\\ Acc.}} & \textbf{\makecell{SST Baseline\\ Debiased Acc.}} & \textbf{\makecell{SST \textit{DeepSoftDebias}\\ Debiased Acc.}} \\
\midrule
BAAI/bge-base-en-v1.5 & Gender & 0.088 & 0.001 & 0.730 & 0.725 & 0.693 \\
BAAI/bge-large-en-v1.5 & & 0.159 & 0.105 & 0.727 & 0.710 & 0.705 \\
google/gemma-2b & & -0.060 & 0.154 & 0.686 & 0.677 & 0.678 \\
google/gemma-7b & & -0.059 & 0.017 & 0.675 & 0.544 & 0.691 \\
GritLM/GritLM-7B & & -0.125 & 0.044 & 0.711 & 0.702 & 0.697 \\
HuggingFaceH4/zephyr-7b-beta & & -0.129 & 0.097 & 0.706 & 0.687 & 0.699 \\
intfloat/multilingual-e5-large-instruct & & -0.037 & 0.096 & 0.729 & 0.720 & 0.724 \\
meta-llama/Llama-2-7b-hf & & 0.009 & -0.032 & 0.701 & 0.692 & 0.686 \\
openai-community/gpt2-large & & 0.042 & -0.038 & 0.664 & 0.665 & 0.669 \\
openai-community/gpt2-xl & & 0.041 & 0.071 & 0.666 & 0.667 & 0.669 \\
tiiuae/falcon-7b & & -0.116 & 0.066 & 0.686 & 0.672 & 0.694 \\
\midrule
BAAI/bge-base-en-v1.5 & Race & 0.094 & 0.092 & 0.730 & 0.709 & 0.683 \\
BAAI/bge-large-en-v1.5 & & 0.104 & 0.099 & 0.727 & 0.727 & 0.695 \\
google/gemma-2b & & -0.041 & 0.164 & 0.686 & 0.665 & 0.686 \\
google/gemma-7b & & -0.055 & 0.133 & 0.675 & 0.549 & 0.678 \\
GritLM/GritLM-7B & & -0.133 & -0.057 & 0.711 & 0.714 & 0.690 \\
HuggingFaceH4/zephyr-7b-beta & & -0.127 & 0.062 & 0.706 & 0.687 & 0.697 \\
intfloat/multilingual-e5-large-instruct & & 0.053 & 0.120 & 0.729 & 0.730 & 0.730 \\
meta-llama/Llama-2-7b-hf & & -0.058 & 0.113 & 0.701 & 0.699 & 0.705 \\
openai-community/gpt2-large & & -0.019 & 0.024 & 0.664 & 0.670 & 0.680 \\
openai-community/gpt2-xl & & 0.149 & 0.180 & 0.666 & 0.665 & 0.692 \\
tiiuae/falcon-7b & & -0.192 & -0.027 & 0.686 & 0.664 & 0.693 \\
\midrule
BAAI/bge-base-en-v1.5 & Religion & 0.054 & 0.078 & 0.730 & 0.716 & 0.694 \\
BAAI/bge-large-en-v1.5 & & 0.153 & 0.175 & 0.727 & 0.718 & 0.697 \\
google/gemma-2b & & 0.118 & 0.278 & 0.686 & 0.679 & 0.682 \\
google/gemma-7b & & 0.127 & 0.194 & 0.675 & 0.548 & 0.685 \\
GritLM/GritLM-7B & & -0.002 & 0.077 & 0.711 & 0.702 & 0.703 \\
HuggingFaceH4/zephyr-7b-beta & & -0.130 & 0.118 & 0.706 & 0.693 & 0.686 \\
intfloat/multilingual-e5-large-instruct & & 0.201 & 0.194 & 0.729 & 0.728 & 0.735 \\
meta-llama/Llama-2-7b-hf & & -0.103 & 0.032 & 0.701 & 0.679 & 0.710 \\
openai-community/gpt2-xl & & 0.247 & 0.251 & 0.666 & 0.671 & 0.679 \\
tiiuae/falcon-7b & & 0.126 & 0.265 & 0.686 & 0.671 & 0.703 \\
\bottomrule
\end{tabular}
\caption{Downstream testing results on Stanford Sentiment Treebank and STS-B Semantic Similarity Dataset. PCC here refers to the Pearson's Coefficient and we report the gain in positive PCC from the Biased embeddings to the debiased embeddings. SST is Stanford Sentiment TreeBank and STS-B is the Semantic Textual Similarity Benchmark}
\label{tab:downstream2}
\end{table*}

\section{Variation of Bias in the Different LLMs}
\label{var}
The presence of biases in  has drawn significant attention from researchers and practitioners. These biases can inadvertently emerge during the training process due to the characteristics of the initial training data. In this study, we explore the factors contributing to bias variation among LLMs, focusing on three prominent models: Llama, Mistral, and Gemma. Our analysis reveals that biases, including those related to gender, race, and culture, are often inherited from the training data. For instance, historical texts may perpetuate gender stereotypes or racial prejudices present in their source material. Llama and Mistral, trained on diverse corpora containing web documents, source code, and mathematical text, exhibit varying degrees of bias. Gemma, released by Google, further demonstrates the impact of training data size, with both 2B and 7B variants drawing from an extensive pool of up to 6 trillion tokens.
\section{Neural Network Schematic}

\begin{figure*}[h]
  \centering
  \includegraphics[width=1\linewidth]{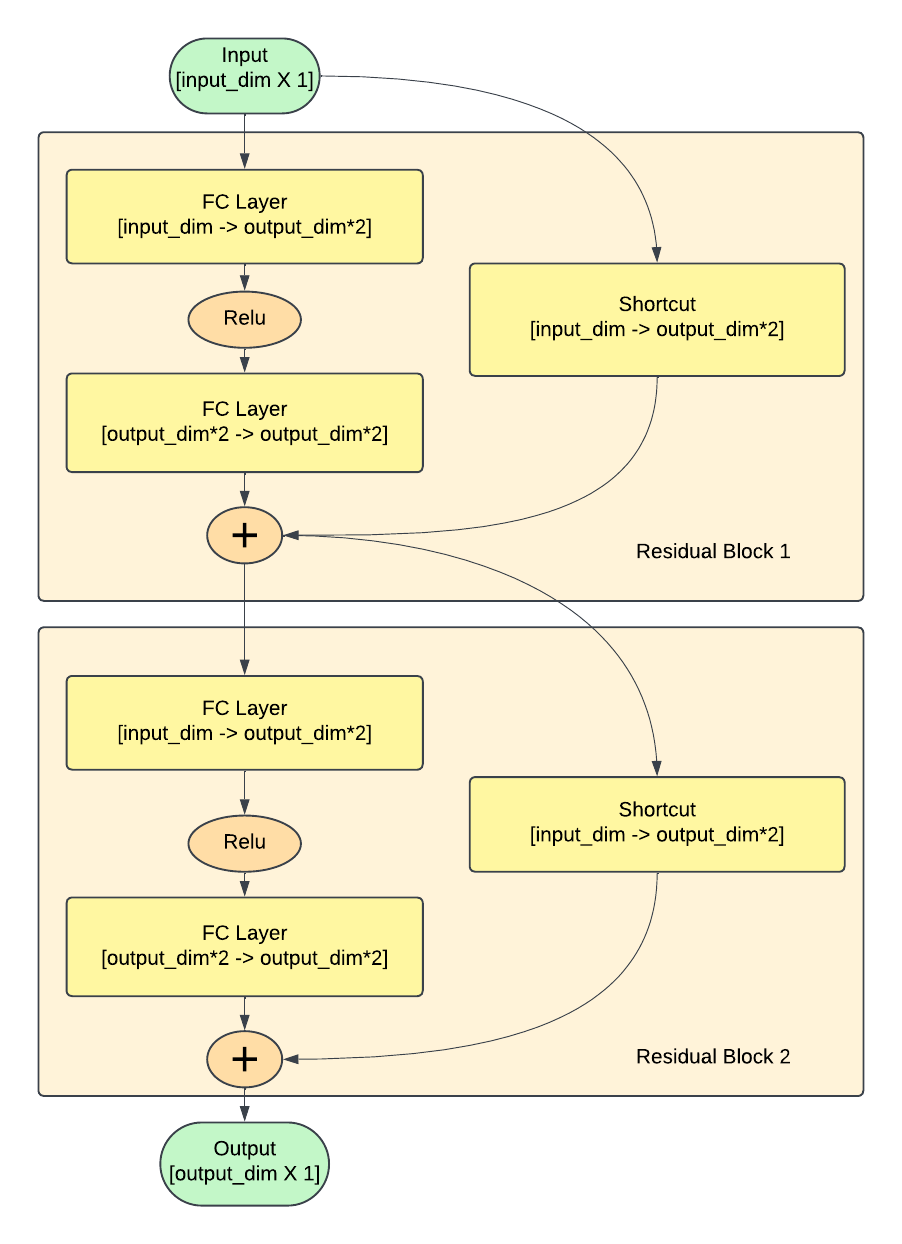} 
  \caption{A schematic of the Embedding Transforming NN with 2 residual blocks}
  \label{fig:NN_schema}
\end{figure*}

\section{Hyperparameters for the different LLMs tested}

\begin{table*}[!h]\centering
\scriptsize
\begin{tabular}{lrrrrrr}\toprule
Model Name &Embedding Dim &Num Res Blocks &LR &Batch Size &Num Epochs \\\midrule
Alibaba-NLP/gte-Qwen2-7B-instruct &3584 &3 &1.00e-5 &1024 &250 \\
BAAI/bge-base-en-v1.5 &768 &1 &5.00e-5 &2048 &100 \\
BAAI/bge-large-en-v1.5 &1024 &1 &5.00e-5 &2048 &100 \\
FacebookAI/roberta-base &768 &1 &5.00e-5 &2048 &100 \\
google-bert/bert-base-uncased &768 &1 &5.00e-5 &2048 &100 \\
google-bert/bert-large-uncased &1024 &1 &5.00e-5 &2048 &100 \\
google/gemma-2-2b &2304 &2 &5.00e-5 &1024 &200 \\
google/gemma-2b &2048 &2 &5.00e-5 &1024 &200 \\
google/gemma-7b &3072 &2 &5.00e-5 &1024 &250 \\
GritLM/GritLM-7B &4096 &3 &5.00e-5 &1024 &300 \\
openai-community/gpt2-xl &1600 &1 &5.00e-5 &2048 &150 \\
Salesforce/SFR-Embedding-2\_R &4096 &3 &1.00e-5 &1024 &300 \\
state-spaces/mamba-1.4b-hf &2048 &2 &5.00e-5 &1024 &200 \\
state-spaces/mamba-2.8b-hf &2560 &2 &5.00e-5 &1024 &250 \\
\bottomrule
\end{tabular}
\caption{Table of the different hyperparameters used with the different LLMs}\label{tab: Hyperparameters}
\end{table*}
\section{Ablation Experiments}
\label{abl}

In our study, we conduct ablation experiments to assess the effectiveness of various debiasing techniques in the realm of natural language processing. These techniques encompassed five distinct scenarios: the utilization of debiased embeddings, the application of the original soft debiasing method, the original debiasing method with the Adam optimizer, \textit{DeepSoftDebias} with the SGD optimizer, and finally, \textit{DeepSoftDebias} with the Adam optimizer. These experiments were gauged based on MAC as the evaluation metric.

Through rigorous experimentation across three biasing directions, we systematically analyze the performance of each method. Our results reveal a consistent trend of incremental improvements as we transitioned from one method to the next. Notably, \textit{DeepSoftDebias}, emerged as the standout performer, boasting the highest mean average cosine similarity score across all evaluated scenarios. In addition, our analysis revealed that substituting the transformation matrix with our neural network approach resulted in the most significant enhancement in the efficacy of the debiasing method. This observation underscores the pivotal role played by neural networks in maximizing the effectiveness of the debiasing techniques. Table \ref{tab:ablation1} presents a visualization of the results of our ablation experiments.

This empirical evidence underscores the robustness and efficacy of our proposed approach in mitigating bias within natural language processing systems. By combining state-of-the-art debiasing techniques with advanced optimization strategies, we have unlocked a powerful methodological framework for enhancing the fairness and accuracy of language models.

\begin{table*}[h]
\centering
\resizebox{0.65\textwidth}{!}{%
\begin{tabular}{lccccc}
\toprule
\textbf{\makecell{Debiasing\\Direction}} & \textbf{Biased} & \textbf{Baseline} & \textbf{\makecell{Baseline\\+ Adam}} & \textbf{\makecell{DeepSoftBias\\+ SGD}} & \textbf{\makecell{DeepSoftBias\\+ Adam}} \\
\midrule
\textbf{Gender} & 0.390 & 0.623 & 0.799 & 0.893 & 0.982 \\
\textbf{Race} & 0.404 & 0.656 & 0.824 & 0.984 & 0.987 \\
\textbf{Religion} & 0.406 & 0.623 & 0.812 & 0.966 & 0.983 \\
\bottomrule
\end{tabular}
}
\caption{Ablations to characterize various design decisions in the development of \textit{DeepSoftDebias}. We start with the transformation matrix, then make incremental additions till we reach the proposed architecture of the \textit{DeepSoftDebias} network.}
\label{tab:ablation1}
\end{table*}

\end{document}